\documentclass[journal]{IEEEtran}
\usepackage{times}
\usepackage{epsfig}
\usepackage{graphicx}
\usepackage{amsmath}
\usepackage{amsthm}
\usepackage{amssymb}
\usepackage{booktabs}
\usepackage{bm}
\usepackage{algorithm}
\usepackage{algorithmic}
\usepackage{mathtools}
\usepackage{enumerate}
\usepackage{paralist}
\usepackage{subcaption}
\usepackage{color}
\usepackage{cite}
\usepackage{colortbl}

\usepackage{caption}
\usepackage{url}            
\usepackage{amsfonts}       
\usepackage{nicefrac}       
\usepackage{microtype}      
\usepackage{threeparttable}
\usepackage{multirow}
\usepackage{graphbox}

\newcommand{\ie}{{\it i.e.}}
\newcommand{\eg}{{\it e.g.}}

\newcommand{\etal}{{\it et al.}}

\begin{document}

\title{Recurrent Exposure Generation for \\ Low-Light Face Detection}


%

\author{
    Jinxiu Liang, Jingwen Wang, Yuhui Quan, Tianyi Chen, Jiaying Liu, Haibin Ling and  Yong Xu*
    \thanks{Jinxiu Liang, Yuhui Quan, Tianyi Chen and Yong Xu are with School of Computer Science and Engineering at South China University of Technology, Guangzhou, China. (Email:~cssherryliang@mail.scut.edu.cn;~csyhquan@scut.edu.cn;~csttychen@mail.scut.edu.cn;~yxu@scut.edu.cn)}
    \thanks{Jingwen Wang is with Tencent AI Lab, Shenzhen, China. (Email: jaywongjaywong@gmail.com)}
    \thanks{Jiaying Liu is with Wangxuan Institute of Computer Technology, Peking University, Beijing, China. (Email: liujiaying@pku.edu.cn)}
    \thanks{Haibin Ling is with the Department of Computer Science, Stony Brook University, Strony Brook, NY, USA. (Email: hling@cs.stonybrook.edu)}
    \thanks{Asterisk indicates the corresponding author.}
}

\maketitle

\begin{abstract}
Face detection from low-light images is challenging due to limited photos and inevitable noise, which, to make the task even harder, are often spatially unevenly distributed.
A natural solution is to borrow the idea from \textit{multi-exposure}, which captures multiple shots to obtain well-exposed images under challenging conditions.
High-quality implementation/approximation of multi-exposure from a single image is however nontrivial.
Fortunately, as shown in this paper, neither is such high-quality necessary since our task is \textit{face detection} rather than \textit{image enhancement}.
Specifically, we propose a novel \textit{Recurrent Exposure Generation (REG)} module and couple it seamlessly with a \textit{Multi-Exposure Detection (MED)} module, and thus significantly improve face detection performance by effectively inhibiting non-uniform illumination and noise issues.
REG produces progressively and efficiently intermediate images corresponding to various exposure settings, and such pseudo-exposures are then fused by MED to detect faces across different lighting conditions.
The proposed method, named \textit{REGDet}, is the first `detection-with-enhancement' framework for low-light face detection.
It not only encourages rich interaction and feature fusion across different illumination levels, but also enables effective end-to-end learning of the REG component to be better tailored for face detection.
Moreover, as clearly shown in our experiments, REG can be flexibly coupled with different face detectors without extra low/normal-light image pairs for training.
We tested REGDet on the DARK FACE low-light face benchmark with thorough ablation study, where REGDet outperforms previous state-of-the-arts by a significant margin, with only negligible extra parameters.
\end{abstract}


\section{Introduction}
\label{sec:intro}

As the cornerstone for many face-related systems, 
face detection has been attracting long-lasting research attention~\cite{yang2002detecting, viola2004robust,jain2010fddb,klare2015pushing,yang2016wider}. 
It has extensive applications in human-centric analysis such as person re-identification~\cite{ding2019feature,huang2019multipseudo} and human parsing~\cite{gong2018instancelevel}. 
Despite great progress in recent decade, face detection remains challenging particularly for images under bad illumination conditions. 
%
Images captured in low-light conditions typically have their brightness reduced and intensity contrast compressed, and thus confuse feature extraction and hurt the performance of face detection. 
Poor illumination also causes annoying noise that further damages the structural information for face detection. 
To make things even worse, the illumination status may spatially vary a lot within a single image. 
For systematic evaluation of face detection algorithms under adverse lighting conditions, a challenging benchmark named DARK FACE~\cite{yang2019ug} is recently constructed, which shows clear performance degradation of state-of-the-art face detectors. For example, DSFD \cite{li2019dsfd} produces an mAP of 15.3\%, in a sharp contrast to above 90\% on the \textit{hard} subset of the popular WIDER FACE~\cite{yang2016wider} benchmark. The dramatic performance degeneration of modern face detectors on the DARK FACE dataset clearly shows that it remains extremely challenging to detect faces under low-light conditions, which is the main focus of this paper.

\begin{figure*}[htbp!]
\centering
    \begin{tabular}{@{}c@{\extracolsep{3pt}}c@{\extracolsep{3pt}}c@{\extracolsep{3pt}}c@{}}
		\includegraphics[width = 0.24\linewidth]{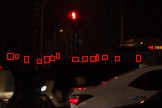} & 
		\includegraphics[width = 0.24\linewidth]{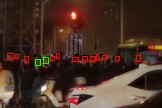} &
		\includegraphics[width = 0.24\linewidth]{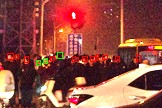} &
		\includegraphics[width = 0.24\linewidth]{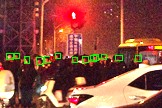} \\
		\includegraphics[width = 0.24\linewidth]{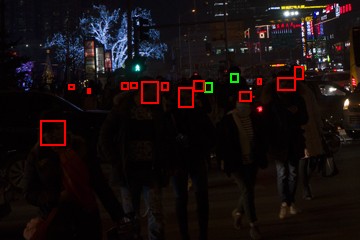} & 
		\includegraphics[width = 0.24\linewidth]{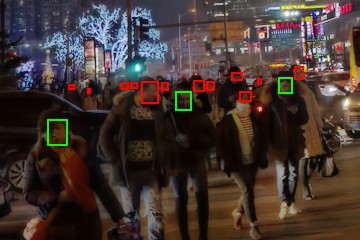} &
		\includegraphics[width = 0.24\linewidth]{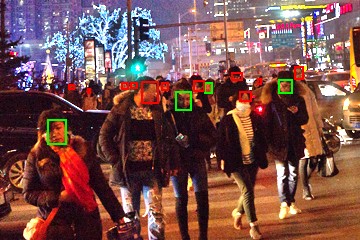} &
		\includegraphics[width = 0.24\linewidth]{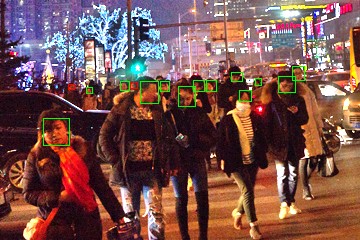} \\
		\includegraphics[width = 0.24\linewidth]{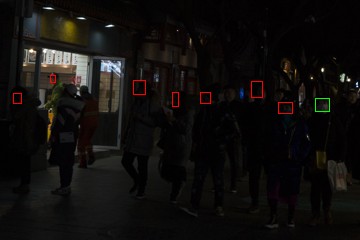} & 
		\includegraphics[width = 0.24\linewidth]{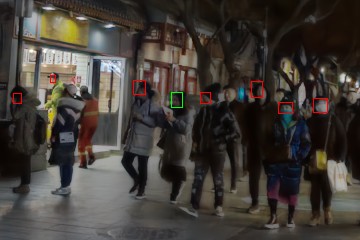} &
		\includegraphics[width = 0.24\linewidth]{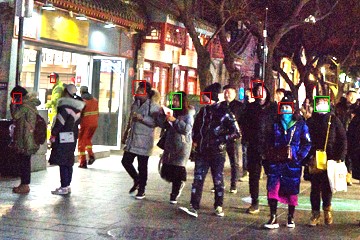} &
		\includegraphics[width = 0.24\linewidth]{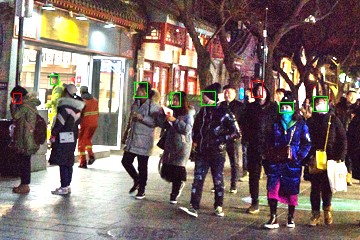} \\
		{\small (a) Low-light image} & {\small (b) KinD~\cite{zhang2019kindling}}  & {\small (c) LIME~\cite{guo2017lime}} & {\small (d) Ours}
	\end{tabular}
\caption{
Detection results of DSFD~\cite{li2019dsfd} on a low-light image (a) and its enhanced versions using KinD~\cite{zhang2019kindling} (b) and LIME~\cite{guo2017lime} (c). \textcolor{green}{Green} and \textcolor{red}{red} boxes indicate true positives and missed targets, respectively. 
It can be seen that the improvement brought by lighting enhancement is very limited. 
By contrast, our result in (d) (plotted on the same image of (c) for better visibility) show clear advantages. 
}
\label{fig:intro}
\end{figure*}


Naturally, one may seek help from low-light image enhancement as preprocessing, as evidenced clearly by the experiments shown in~\cite{yang2019ug}. 
However, as illustrated in Fig.~\ref{fig:intro} (b-c), there is still a large room for improvement. 
For one reason, image enhancement aims to improve visual/perceptual quality for the entire image, which is not fully aligned with the goal of face detection. 
For example, the smoothing operations for enhancing noisy images could compromise the feature discriminability that is critical for detection.
This suggests a close integration between the enhancement and detection components, and points to an end-to-end `detection-with-enhancement' solution. 

Another reason lies in that the illumination in the original image may vary greatly in different regions. 
Consequently, it is hard to expect a single light-enhanced image to handle well facial regions under different lighting conditions in terms of detection. 
This suggests the use of a multiple enhancement strategy and brings our attention to the \textit{multi-exposure} technique. 
In particular, when it is difficult to obtain a well-exposed image with a single shot, the technique takes multiple shots with varying camera settings. 
Such multi-exposure images are then fused for light enhancement.
Similarly and intuitively, we may generate multi-exposure images and then detect faces from them to cover different exposure conditions. 
However, automatically deriving high quality multi-exposure images from a single image is nontrivial~\cite{wang2015pseudomultipleexposurebased}, let alone a low-light image -- but such high quality is not required for \textit{face detection}. 
It is the mechanism for capturing information at different exposures that matters.


Driven by the above motivations, we propose a novel end-to-end low-light face detection algorithm named \emph{REGDet}. 
REGDet contains two sequentially connected modules, a \textit{Recurrent Exposure Generation} (REG) module and a \textit{Multi-Exposure Detection} (MED) module. 
From an input image, REG generates a sequence of pseudo-exposures to loosely mimic the effect of the highly non-linear process of in-camera multi-exposure. 
This is done by assembling a set of ConvGRUs marching in two directions: one direction points progressively and recurrently to the degree of exposure, while the other guides encoder-decoder structures to produce exposure compensated images. 
Then, these pseudo-exposures are fed into MED, which adapts generic face detectors so as to fuse `multi-exposure' information of different pseudo-exposures smoothly. 
With the two modules collaborated together, REGDet not only encourages rich interaction and feature fusion across different illumination levels, but also enables end-to-end learning of effective low-light processing tailored for face detection. 
Moreover, as shown in our experiments, REG can be flexibly coupled with different face detectors without extra low/normal-light image pairs. 
We tested REGDet on the DARK FACE low-light face benchmark with thorough ablation study. In the experiments, REGDet outperforms previous state-of-the-arts by a significant margin, with only negligible extra parameters.

To summarize, we make the following contributions:
\begin{itemize}
\item The first end-to-end `detection-with-enhancement' solution, REGDet, for face detection under poor lighting conditions, 
\item A novel and lightweight recurrent exposure generation module to tackle the non-uniform darkness issue, 
\item A flexible framework compatible to existing face detectors, 
\item New state-of-the-art performance on the publicly available benchmark.
\end{itemize}


\section{Related Work}
\label{sec:related}

The focus in this paper is on developing a learning solution for low-light face detection. In the following we describe previous studies from three aspects: low-light image enhancement, low-light face detection, and gated recurrent networks. 

\subsection{Low-Light Image Enhancement}

Low-light image enhancement has been a popular topic recently for improving the perceptual quality of images. Early solutions often rely on local statistics or intensity mapping, \eg, histogram equalization~\cite{arici2009histogram} and gamma correction~\cite{farid2001blind}. 
Later solutions are often based on the Retinex theory~\cite{land1977retinex} which assumes an image as a combination of a reflectance map that reflects the physical characteristic of scene objects and a spatially smooth illumination map. 
Thus developed solutions 
focus on resolving the ambiguity between illumination and reflectance by imposing certain priors on a variational model based on empirical observations 
(\eg, \cite{wang2013naturalness,fu2016fusionbased,guo2017lime,fu2016weighted,li2018structurerevealing}).
More recently, deep learning-based solutions boost further the image enhancement quality.
These recent methods often produce impressive results for enhancing low-light images (\eg,~\cite{wang2019underexposed,wang2018gladnet, wei2018deep, zhang2019kindling}). 
However, the performance gain, when applied to low-light face detection, is still far from saturated~\cite{yang2019ug}. 
As discussed in previous section, this is partly due to their different goal with face detection, dealing with uneven illumination inside a single image, and weak collaboration with a face detection module. 


The most related work to ours in low-light image enhancement is the multi-exposure fusion-based method BIMEF~\cite{ying2017bioinspired}. BIMEF first synthesizes a brighter image by a Brightness Transform Function (BTF) with fixed camera parameters, and then blends it with the original low-light image into a better one. Our method shares the idea of generating multi-exposure images, but is driven by a very different goal, \ie, face detection. Consequently our model is learned end-to-end for the goal. Moreover, BIMEF does not consider the inevitable noise in low-light images and does not leverage the powerful data-driven modeling capacity of deep learning. 

\subsection{Low-Light Face Detection}
\label{sec:related_detection}

With the advent of large-scale face detection datasets~\cite{jain2010fddb,klare2015pushing,yang2016wider} and the proliferation of deep learning technologies~\cite{girshick2015fast,ren2015faster,liu2016ssd,lin2017feature}, 
face detection in unconstrained environments (a.k.a. `in the wild') has made remarkable progress~\cite{najibi2017ssh,hao2017scaleaware,hu2017finding,zhang2017s3fd,shi2018realtime,tang2018pyramidbox,najibi2019farpn,li2019dsfd}. 
Most recent technological developments have focused on robustness to geometric variance.
Typical geometric distortion includes scale variation, deformation, occlusion and so on. 
For scale variation, researchers have proposed many effective strategies based on the idea of multi-scale analysis:
designing image pyramids with different image scales~\cite{hu2017finding}, designing a pre-defined set of anchor boxes with different sizes and aspect ratios~\cite{jiang2017face,ming2019group,najibi2019farpn}, detecting at different layers of the network~\cite{najibi2017ssh,zhang2017s3fd} and so on.
Deformable part-based model improves deformation invariance by decomposing the task of face detection into detecting different facial parts~\cite{yang2015facial}. The idea of face calibration is explored to obtain deformation invariance in~\cite{shi2018realtime}. 
Spatial context aggregation is a modern strategy for obtaining invariant features. Existing context aggregation techniques include enlarging receptive field by dilated convolution~\cite{chi2019selective}, multi-layer fusion~\cite{sun2018face} and top-down feature fusion~\cite{li2019dsfd,tang2018pyramidbox}.

Low-light face detection has been attracting research attention for a long time. 
In the era of hand-crafted features, enduring efforts have been made to understand and handle the non-uniform illumination issue~\cite{han2013comparative, shengyeyan2008locally,levi2004learning}. 
Recently, there are increasing interests in data-driven approaches for face detection on low-quality images such as low-resolution images and low-light images~\cite{zhou2018survey, nada2018pushing,yang2019ug}. Illumination variation is known to be a major challenge for modern face detection algorithms~\cite{adini1997face, zhou2018survey}. Pioneering approaches preprocess images by intensity mapping such as logarithmic transform~\cite{adini1997face} and gamma transform~\cite{shan2003illumination}. 
Photometric normalization is another commonly adopted method that counteracts the varying lighting conditions in hand-crafted feature~\cite{shengyeyan2008locally,chen2006illumination} and deep learning-based methods~\cite{zhou2018survey,liu2016ssd}. 
Hand-crafted feature based methods derive the illumination invariance from various priors such as image differences or gradients~\cite{adini1997face,han2013comparative}, while deep learning-based methods use random photometric distortions as augmentation to implicitly enhance the illumination invariance~\cite{zhang2017s3fd,tang2018pyramidbox,li2019dsfd}. 
Despite previous studies, face detection in extremely adverse light conditions has been under explored, due partly to the lack of high quality labeled data. Addressing this issue, Yang \etal~present a large manually labeled low-light face detection dataset, DARK FACE, and show that existing face detectors perform poorly on the task~\cite{yang2019ug}. Our work is thus motivated and evaluated on the benchmark, and outperforms clearly previous arts.  
Baseline experiments have shown that, despite of the outstanding success achieved nowadays, even the best well-trained face detectors are less than ideal if the images are simply pre-processed using existing low-light enhancement methods~\cite{yang2019ug}. 

\subsection{Gated Recurrent Networks}

Gated Recurrent Networks are the most related work to ours from the learning aspect. 
Gated recurrent unit (GRU) in recurrent networks is a gating mechanism to adaptively control how much each unit remembers or forgets for sequence modeling~\cite{cho2014learning}. 
It was first proposed and applied to task of machine translation. 
ConvGRU~\cite{ballas2016delving} extends the fully-connected layers in GRU with convolution operations to model correlations among image sequence. 
The design of the REG module is greatly inspired by~\cite{li2018recurrent}. However, 
the learning of the REG module is performed with a proposed pseudo-supervised pre-training strategy and the implicit guidance of a follow-up detection module instead of ground-truth data.
Moreover, instead of predicting rain streak layer by residual learning, REB directly learns to generate various pseudo-exposures.

\section{The Proposed Method}
\label{sec:motivation}

\begin{figure*}[htbp!]
    \centering
     \includegraphics [width=1.0\linewidth]{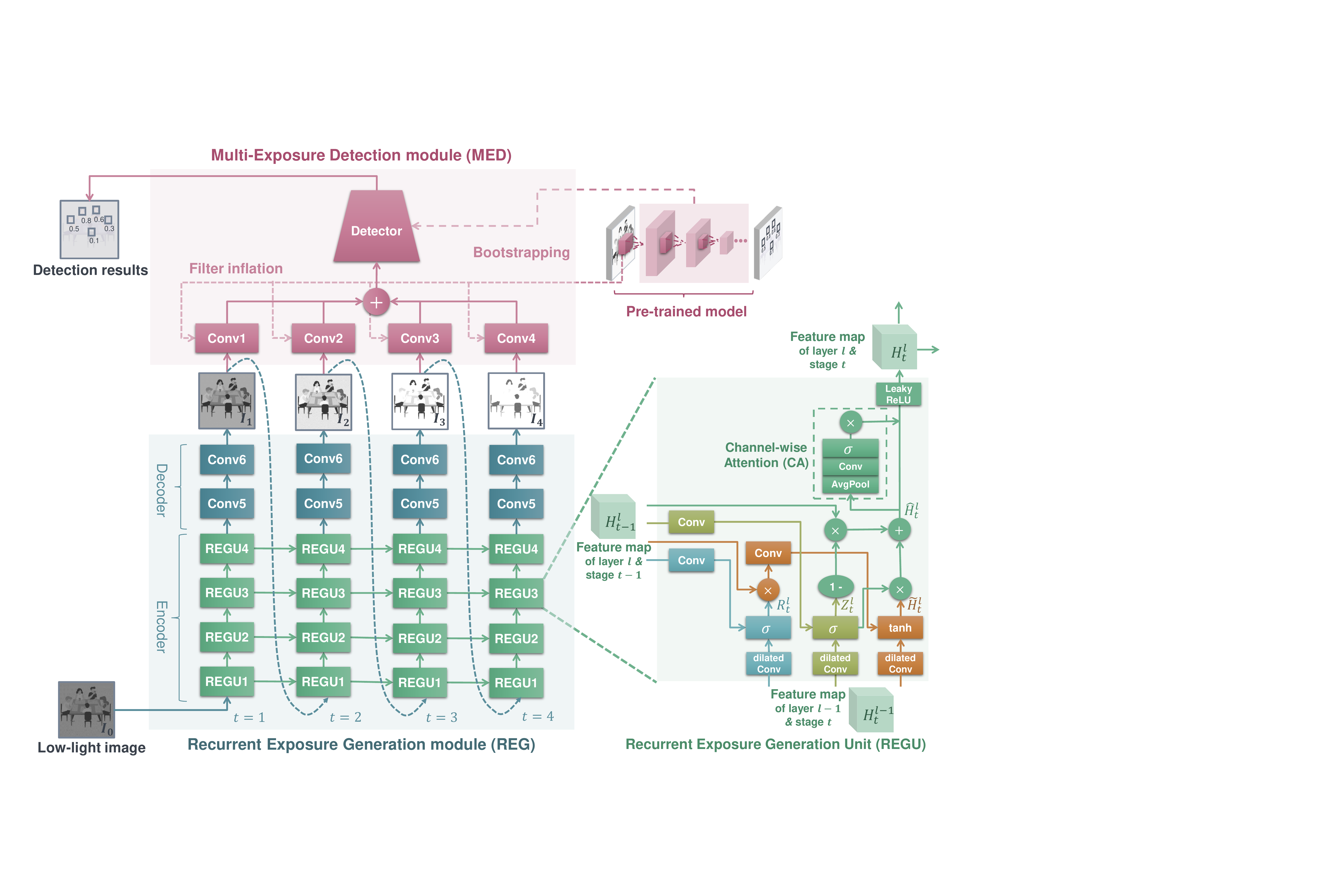}
    \caption{
    The main framework of the proposed REGDet for low-light face detection.
    }
    \label{fig:method}
\end{figure*}

As shown in Fig.~\ref{fig:method}, the proposed REGDet involves two main modules, the Recurrent Exposure Generation module (REG) and the Multi-Exposure Detection module (MED). 
To loosely mimic the complex and highly non-linear in-camera multi-exposure process,  
REG generates progressively brighter images while encoding historical regional information. 
These pseudo-exposures are then fed into MED to produce face bounding boxes. 
The two modules are coupled together to form an end-to-end framework.

\subsection{The Recurrent Exposure Generation Module}
\label{sec:method}

To progressively generate $T$ pseudo-exposures from a low-light input image $I_0$, a natural solution is to generate the next image $I_{t+1}$ by an NN conditioned on the previous image $I_{t}$. 
However, as there exists non-uniform darkness in low-light images, such strategy could lead to locally over-smoothed or over-exposed regions, and consequently hurt the face detection task that relies seriously on discriminative details. 

To address the above issue, the proposed Recurrent Exposure Generation (REG) module leverages historical generated images to maintain critical region details in a Recurrent Neural Network (RNN) framework. 
%
Starting from $I_0$ and initial hidden state $H_0=\mathbf{0}$, REG generates recurrently $T$
intermediate pseudo-exposures $\mathbb{I}=\{I_t\}_{t=1}^T$ formulated as
\begin{equation}
 (I_{t}, H_{t}) = \mathcal{G}_{\omega}(\mathcal{F}_{\theta}(I_{t-1}, H_{t-1})), \ \ t=1, 2, \ldots, T,
\end{equation}
where $\mathcal{F}_\theta$ and $\mathcal{G}_\omega$ denote the encoder and the decoder of the proposed module, respectively, with corresponding parameters $\theta$ and $\omega$. 
The encoder consisting of four cascaded convolutional recurrent layers is responsible for transforming the input image into features maps of multiple scales (layers), 
while the decoder consisting of two convolutional layers learns to decode the feature maps back to images, as shown in Fig.~\ref{fig:method}.

At stage $t>0$, $H_t={\{H_t^l\}}_{l=1}^{L}$ where $H_t^l$ denotes feature map from the $l$-th layer. Initialized by $H^0_t=I_{t-1}$, the feature maps are produced by our recurrent exposure generation unit (REGU) $\mathcal{F}^l$ as
\begin{equation}
{H}_t^l = \mathcal{F}^l (H_t^{l-1}, H_{t-1}^l), \ \ l=1, 2, ..., L.
\end{equation}
In particular, REGU is designed based on the Convolutional Gated Recurrent Unit (ConvGRU)~\cite{ballas2016delving} for performance and memory consideration, as shown in the right part of Fig.~\ref{fig:method}. 
An REGU $\mathcal{F}^l$ in the $l$-th layer can be described by the following equations:
\begin{align}
Z_t^l & =\sigma\left( W_z^l * H_t^{l-1} + U_z^l * H_{t-1}^l \right),\\
R_t^l & =\sigma\left( W_r^l * H_t^{l-1} + U_r^l * H_{t-1}^l \right),\\ 
{\widetilde{H}}_t^l & = \operatorname{tanh} \left( W_h^l * H_t^{l-1} + U_h^l * (R_t^l \odot H_{t-1}^l) \right),\\
{\hat{H}}_t^l & =(1-Z_t^l) \odot H_{t-1}^l + Z_t^l \odot  {\widetilde{H}}_t^l,\\
{H}_t^l & =  \xi(\mathcal{A}^l({\hat{H}}_t^l)),
\end{align}
where $Z$ and $R$ are update and reset gates, respectively, which decide the degree to which the unit updates or resets its historical encoding information, $\sigma(x) = \frac{1}{1+e^{-x}}$ is sigmoid function, $\odot$ denotes the Hadamard product, $*$ denotes a convolution operator, filters $W$ and $U$ are for dilated and regular convolution respectively.
$\xi$ denotes leaky ReLU~\cite{maas2013rectifier} activation function
\begin{equation}
\xi(x)=
    \begin{cases}
    \alpha x, & x<0,
    \\
    x, & x\ge 0,
    \end{cases}
\end{equation}
where $\alpha=0.2$ denotes the negative slope.
Given a feature map $H\in\mathbb{R}^{X\times Y\times C}$, the channel-wise attention (CA)~\cite{wang2019ecanet}  $\mathcal{A}^l$ can be computed as
\begin{equation}
    \mathcal{A}^l(H) = \mathcal{A}_s\left( \sigma(W_a^l * \mathcal{A}_g(H)), H \right),
\end{equation}
where $\mathcal{A}_g(H)=\frac{1}{XY}\sum_{i=1,j=1}^{X,Y}H_{ij}$ is channel-wise global average pooling, $W_a^l$ denotes a 1D  convolution kernel with kernel size 3 and $\mathcal{A}_s$ denotes channel-wise multiplication between the feature map and the obtained channel weighting vector.

REGB has several extensions compared with the standard ConvGRU. First, an important component in our REGU is the channel-wise attention, which is integrated in each unit before activation except for the last one. Like in other vision tasks~\cite{wang2019ecanet}, such an efficient mechanism enables appropriate cross-channel interaction inside a feature map and therefore helps aggregate spatial global information and recalibrate the feature map at each step. Second, REGU uses leaky ReLU~\cite{maas2013rectifier} as the activation function to alleviate the `dying ReLU' issue,~\ie~, some neurons going through the flat side of zero slope stop being updated.
Third, to tackle the issue of unevenly distributed darkness, different dilation rates ($2^l$ in the $l$-th layer) are used in different convolutional layers of the encoder to obtain progressively larger receptive fields while maintaining small parameter cost. 

\subsection{Pseudo-Supervised Pre-Training of the REG Module}
\label{sec:reg}
To enable good diversity and complementarity of the generated sequence, we adopt a pseudo-supervised pre-training strategy which leverages pseudo ground-truth images corresponding to different exposures. 
The pseudo ground-truth images $\{\hat{I}_t\}_{t=1}^{T}$ are generated from $I_0$ by a camera response model~\cite{ying2017bioinspired} that characterises the relationship between pixel values and exposure ratios. 
A camera response model contains a camera response function (CRF), \ie, the nonlinear function relating camera sensor irradiance with image pixel value, and a brightness transform function (BTF), \ie, the mapping function between two images captured in the same scene with different exposures~\cite{ren2019lecarm}.
Once the parameters of CRF corresponding to a specific camera is known, the parameters of BTF can be estimated by solving the comparametric equation~\cite{mann2000comparametric}. 
However, the information about the cameras to estimate accurate camera response models is often far from enough in the publicly available low-light face detection dataset. 
Therefore, we adopt the camera response model proposed in~\cite{ying2017bioinspired} that can characterize a general relationship between the pixel values and exposure ratios when no camera information is available.
Its BTF is in the form of Beta-Gamma Correction
\begin{equation}
    \mathcal{B}(P,k)=e^{b(1-k^a)}P^{(k^a)},
\end{equation}
where $P$ and $k$ denote the pixel value and the exposure ratio respectively, and the camera parameters $a=-0.3293,b=1.1258$ are estimated by fitting the 201 real-world camera response curves in the DoRF database~\cite{grossberg2004modeling}. 
Specifically, the exposure ratios are $k,\dots,k^T$, where the base ratio is empirically set as $k=2.4$.

The REG module is then guided to generate images corresponding to diversified exposures. 
To measure the distance between the generated image $I_t$ and the pseudo ground-truth $\hat{I}_t$ produced from $I_0$ with parameter $k^t$, we use a combination of $\ell_1$ norm and the Structure Similarity (SSIM) index~\cite{wang2004image} that reflects the difference on luminance and contrast,  which is formulated as
\begin{equation}
    \mathcal{L}_\mathrm{reg}(I, \hat{I})=\frac{1}{TN}\sum_t{ \sum_{p} {(\lVert I_t - \hat{I}_t \rVert_1 + 1-\operatorname{SSIM})}},
\end{equation}
and the SSIM meature is defined as 
\begin{equation}
    \operatorname{SSIM} = {\frac{(2\mu_{p_t} \mu_{\hat{p}_t}+C_1)(2 \sigma_{{\hat{p}_t}{p_t}}+C_2)}{(\mu_{p_t}^2+\mu_{\hat{p}_t}^2+C_1)(\sigma_{p_t}^2+\sigma_{\hat{p}_t}^2+C_2)}},
\end{equation}
where means $\mu$ and deviations $\sigma$ are computed by applying a Gaussian filter at pixel $p_t$ of image $I_t$ and $N$ denotes the number of pixels in the image.
Following common practice in image enhancement, we randomly crop $64 \times 64$ patches followed by random mirror, resize and rotation for data augmentation.

As the pseudo ground-truth images have inevitable noise and artifacts, we adopt the early stopping strategy to prevent over-fitting to those noise and artifacts. 
Specifically, the pre-training stops when the average PSNR of  ${I}_t$ compared to  $\hat{I}_t$ reaches around 25. 
We use the training split of the DARK FACE dataset to perform the pseudo-supervised pre-training. 
As our method does not rely on any external low/normal-light image pairs, it enjoys good scalability and can be fairly compared to other approaches. 
This pre-training practice can be expected to speedup the joint training process and boost the final detection performance. The performance comparison can be found in Table \ref{tab:ablation_pretraining}.

\begin{figure*}[htbp!]
\centering
        \begingroup
        \renewcommand{\arraystretch}{0} 
        \begin{tabular}{@{}c@{\extracolsep{3pt}}c@{\extracolsep{3pt}}c@{\extracolsep{3pt}}c@{\extracolsep{3pt}}c@{}}
             \includegraphics [width=0.19\linewidth]{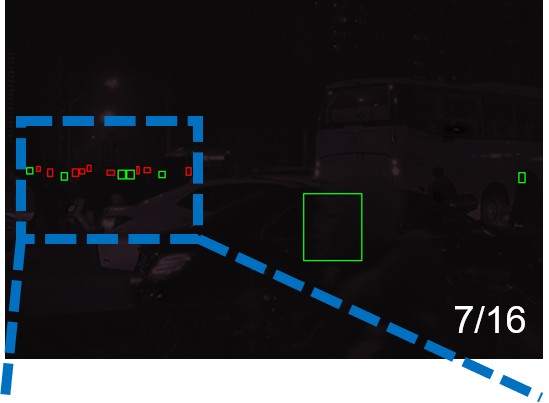} &  
             \includegraphics [width=0.19\linewidth]{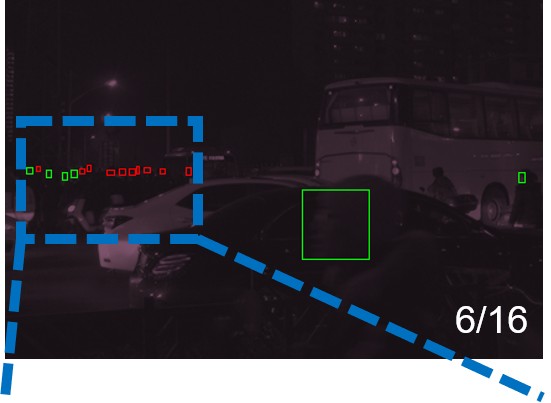} &
             \includegraphics [width=0.19\linewidth]{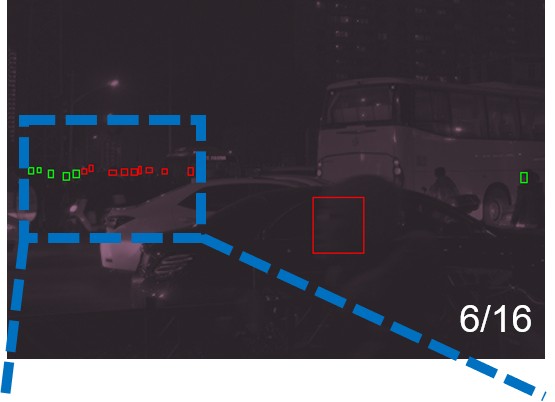} &  
             \includegraphics [width=0.19\linewidth]{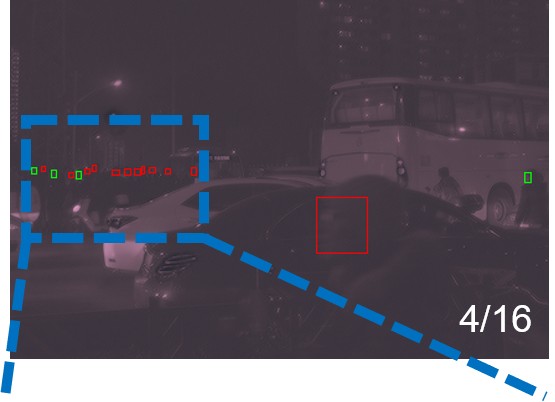} &
             \includegraphics [width=0.19\linewidth]{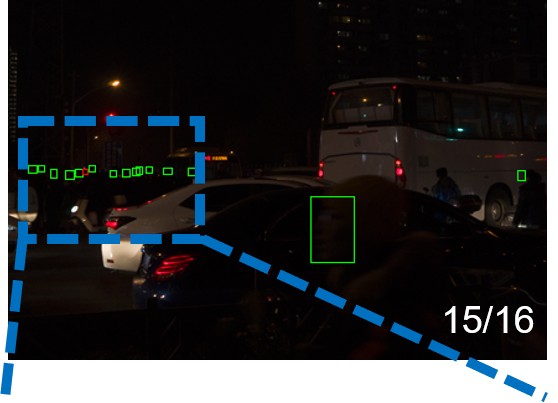}\\
             \includegraphics [width=0.19\linewidth]{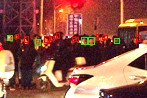} &  
             \includegraphics [width=0.19\linewidth]{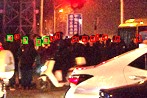} &
             \includegraphics [width=0.19\linewidth]{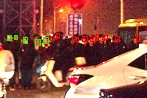} &  
             \includegraphics [width=0.19\linewidth]{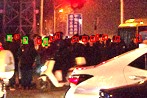} &
             \includegraphics [width=0.19\linewidth]{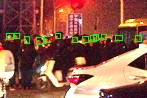}\\[10pt]
             \includegraphics [width=0.19\linewidth]{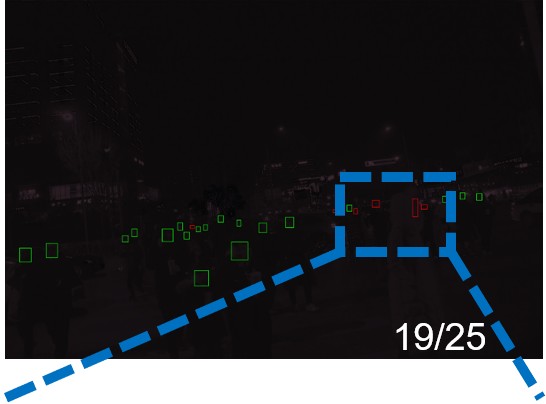} &  
             \includegraphics [width=0.19\linewidth]{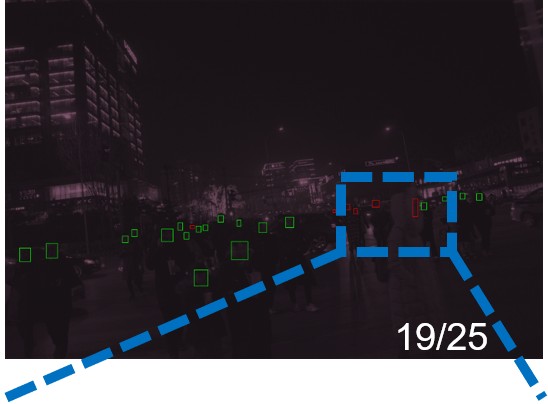} &
             \includegraphics [width=0.19\linewidth]{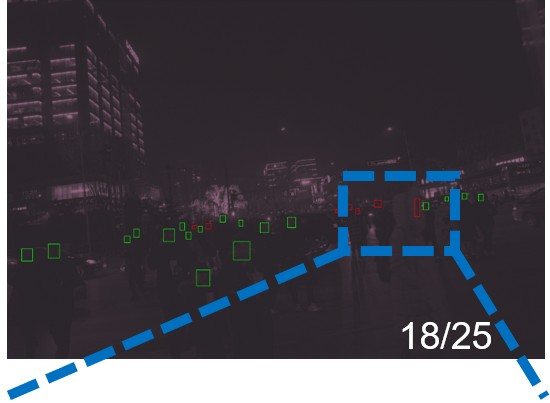} &  
             \includegraphics [width=0.19\linewidth]{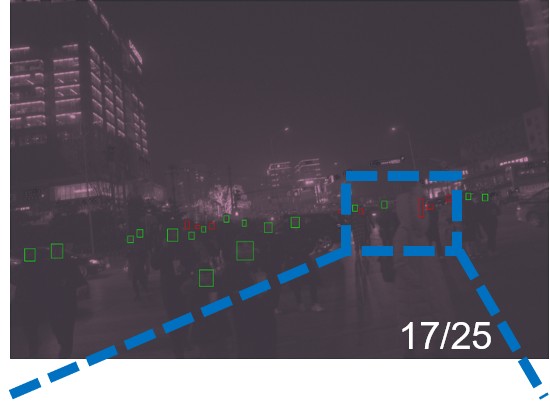} &
             \includegraphics [width=0.19\linewidth]{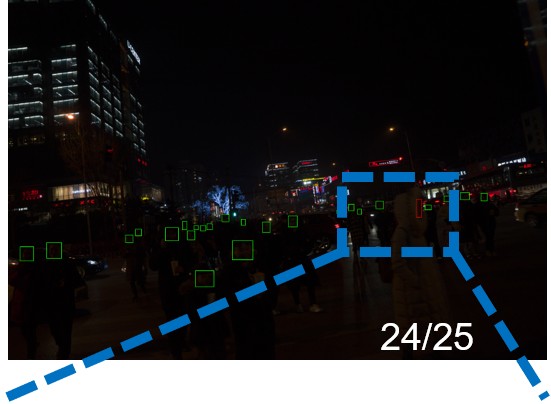}\\
             \includegraphics [width=0.19\linewidth]{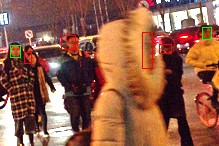} &  
             \includegraphics [width=0.19\linewidth]{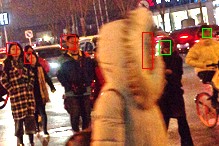} &
             \includegraphics [width=0.19\linewidth]{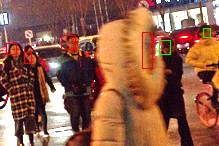} &  
             \includegraphics [width=0.19\linewidth]{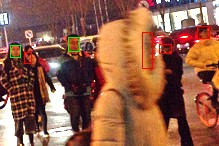} &
             \includegraphics [width=0.19\linewidth]{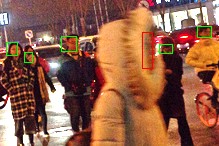}\\[10pt]
             \includegraphics [width=0.19\linewidth]{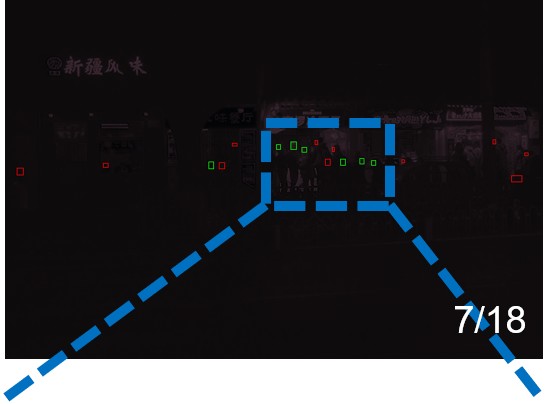} &  
             \includegraphics [width=0.19\linewidth]{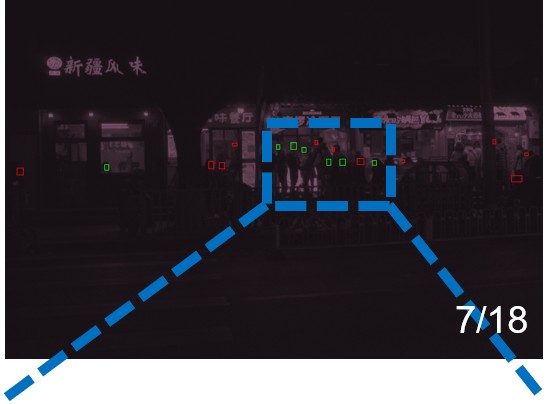} &
             \includegraphics [width=0.19\linewidth]{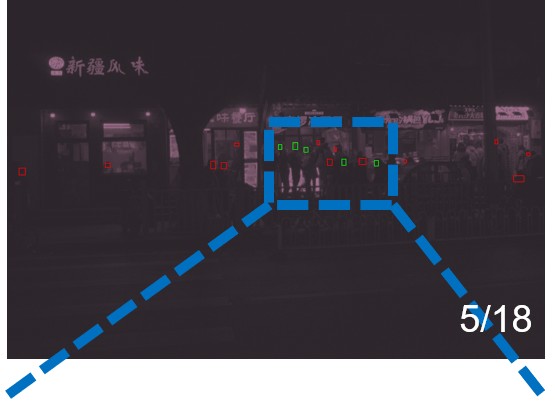} &  
             \includegraphics [width=0.19\linewidth]{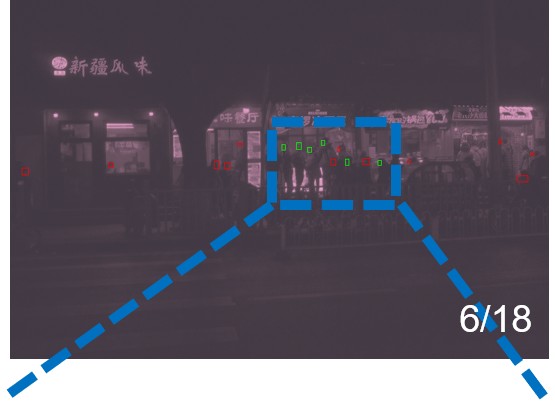} &
             \includegraphics [width=0.19\linewidth]{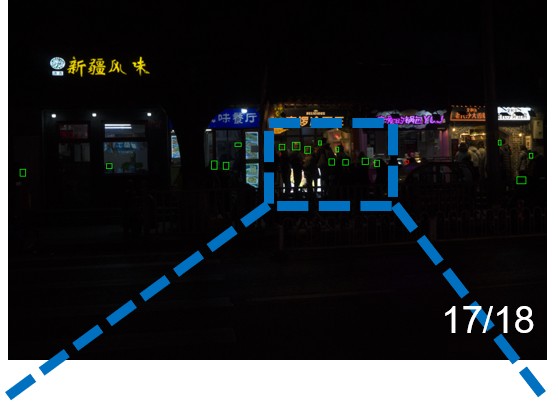}\\
             \includegraphics [width=0.19\linewidth]{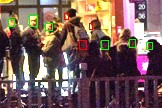} &  
             \includegraphics [width=0.19\linewidth]{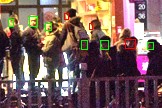} &
             \includegraphics [width=0.19\linewidth]{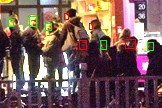} &  
             \includegraphics [width=0.19\linewidth]{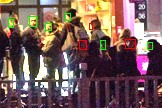} &
             \includegraphics [width=0.19\linewidth]{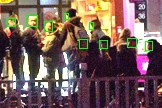}\\[8pt]
             (a) Detect on $I_1$ & (b) Detect on $I_2$ & (c) Detect on $I_3$ & (d) Detect on $I_4$ & (e) REGDet \\
         \end{tabular}
         \endgroup
    \caption{
    The left four are detection results on intermediate $I_1, I_2, I_3$ and $I_4$ generated from the REG module, which show complementarity among the generated images, supporting the effectiveness of our proposed REG module. {Note these `images' are linearly normalized for visualization so that the minimum (maximum) value corresponds to 0 (255).}
    The rightmost column shows our final detection result, where more faces (14 out of 15) are successfully localized, showing superiority of the proposed MED module.
    \textcolor{green}{Green} and \textcolor{red}{red} boxes indicate true positives and missed targets, respectively. The zoom-in versions on the second row are enhanced by LIME~\cite{guo2017lime} for better visibility.
    }
    \label{fig:visual}
\end{figure*}

To understand and verify the complementarity of the generated sequence from the REG module, we visualize them in Fig.~\ref{fig:visual}. The detection results on the generated images using the pre-trained DSFD detector in the left four images show good \textit{complementarity} between different generated images, indicating that the REGDet learns to generate a complementary detection-oriented image sequence to benefit subsequent face detection.

\subsection{The Multi-Exposure Detection Module}
\label{sec:med}

Once the multiple pseudo-exposures $\mathbb{I}$ are created by the REG module, a straightforward strategy is to separately feed them into a face detector and fuse their corresponding detected bounding boxes, \ie, \textit{late fusion}. 
This is however computationally expensive as it requires multiple runs of the detection process. 
Instead, we introduce a resource efficient strategy to fuse the low-level features extracted from $\mathbb{I}$ in early stage of detection. Such strategy not only takes advantage of available pre-trained face detectors, but also allows the collaboration among different pseudo-exposures. 


Specifically, the proposed Multi-Exposure Detector (MED) module integrates a generic pre-trained CNN-based face detection algorithm, named \textit{base detector} with \textit{early fusion}. We tailor its first convolutional layer using \textit{filter inflation} technique~\cite{carreira2017quo} in the channel dimension so that the detector can simultaneously process multiple images and perform adaptive integration, as shown in Fig.~\ref{fig:method}. The weights of the $T$ convolutional layers are bootstrapped from the first layer in the pre-trained base detector, by duplicating and normalizing the pre-trained filter weights $T$ times, which helps maintain better discriminative and complementary regional clues across different pseudo-exposures.
%
Formally, MED $\mathcal{M}$ simultaneously predicts the confidences ${p}=\{p_i\}_{i=1}^{N_a}$ and the bounding box coordinates ${g}=\{g_i\}_{i=1}^{N_a}$ of anchor boxes indexed by $1,2, \dots, N_a$ as
\begin{equation}
    ({p}, {g}) = \mathcal{M} \left(\mathbb{I}\right) ,
\end{equation}
where $N_a$ denotes the number of anchors, $p_i$ measures how confident the $i$-th anchor is a face and $g_i$ is a vector representing the $4$ parameterized coordinates of the predicted face boxes.
Following~\cite{liu2016ssd}, we use weighted sum of the confidence loss and the localization loss:
\begin{equation}
    \mathcal{L}\left({p}, \hat{p}, {g}, \hat{g}\right) = \frac{1}{N_a} \sum_i \mathcal{L}_\mathrm{conf}\left(p_i, \hat{p}_i\right) + \frac{\lambda}{N_p} \sum_i \hat{p}_i \mathcal{L}_\mathrm{loc}\left(g_i, \hat{g}_i\right),
\end{equation}
where $N_p$ denotes the number of positive anchors, $\lambda$ is used to balance the two loss terms, 
the ground-truth label $\hat{p}_i$ represents whether the $i$-th anchor is positive (a.k.a., is a face), and $\hat{g}_i$ is the ground-truth bounding box assigned to the anchor. 
The confidence (classification) loss $\mathcal{L}_\mathrm{conf}\left(p_i, \hat{p}_i\right)$ is a two-class (face or background) softmax loss, 
\begin{equation}
    \mathcal{L}_\mathrm{conf}\left(p_i, \hat{p}_i\right) = \hat{p}_i \log \left(p_i\right) + (1 - \hat{p}_i) \log \left(1 - p_i\right),
\end{equation}
where the $\hat{p}_i$ in the second term means that the localization loss is only calculated for those positive anchors. 
Following \cite{girshick2015fast}, the localization loss $\mathcal{L}_\mathrm{loc}\left(g_i, \hat{g}_i\right)$ is defined as the smooth $\ell_1$ loss, \ie, the distance between the predicted box $g_i$ and the ground-truth $\hat{g}_i$  measured by Huber norm
\begin{equation}
    \mathcal{L}_\mathrm{loc}\left(g_i, \hat{g}_i\right) = \sum_{j \in \{x,y,h,w\}} \mathcal{H} \left( g_i^{(j)} - \hat{g}_i^{(j)} \right),
\end{equation}
 where the Huber norm $\mathcal{H}(\cdot )$ is defined as
\begin{equation}
    \mathcal{H} \left( x  \right) = \left\{ 
    \begin{array}{cc}
    0.5 x^2     & \operatorname{if} |x| < 1 \\
    |x| - 0.5   & \operatorname{otherwise}.
    \end{array}{}
    \right .
\end{equation}
The Huber norm is less sensitive to outliers than the $\ell_2$ norm.

Being an end-to-end system, REGDet allows joint optimization of the REG and MED modules during learning. Intuitively, MED provides facial location information to guide REG such that the facial regions could be specially enhanced for the purpose of detection. An example detection result is shown in the rightmost column of Fig.~\ref{fig:visual}, and it shows that REGDet successfully localizes far more faces than simply applying the base detector on different intermediate images.

It is worth noting that MED is flexible in choosing the base detector. In our experiments, several state-of-the-art algorithms such as DSFD \cite{li2019dsfd}, PyramidBox \cite{tang2018pyramidbox} and S3FD \cite{zhang2017s3fd} all demonstrate clear performance improvement when embedded in REGDet.

\begin{figure}[htbp!]
    \centering
    \begin{tabular}{@{}c@{}}
        \includegraphics [width=1.0\linewidth]{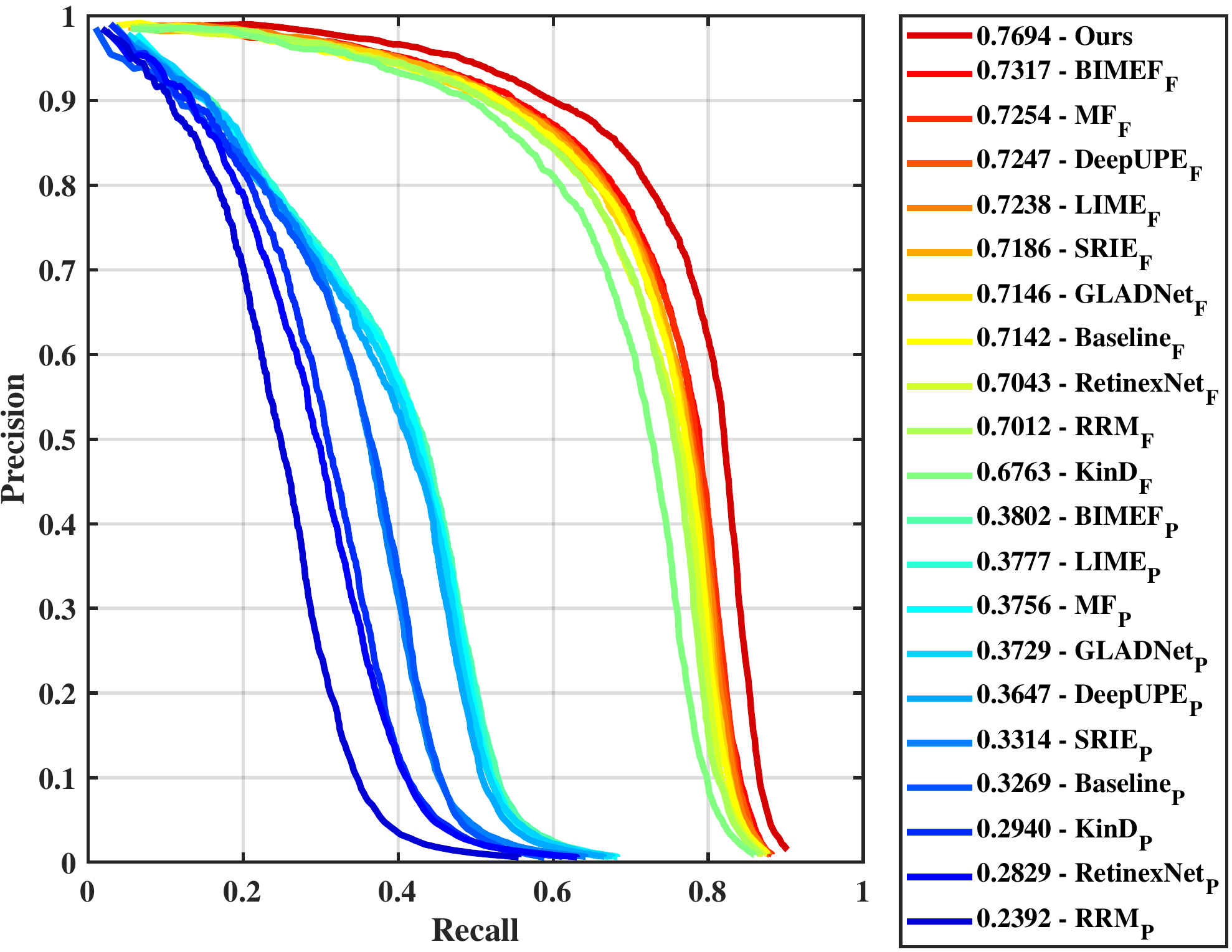}\\
        {\small (a) DSFD~\cite{li2019dsfd}} \\
        \includegraphics [width=1.0\linewidth]{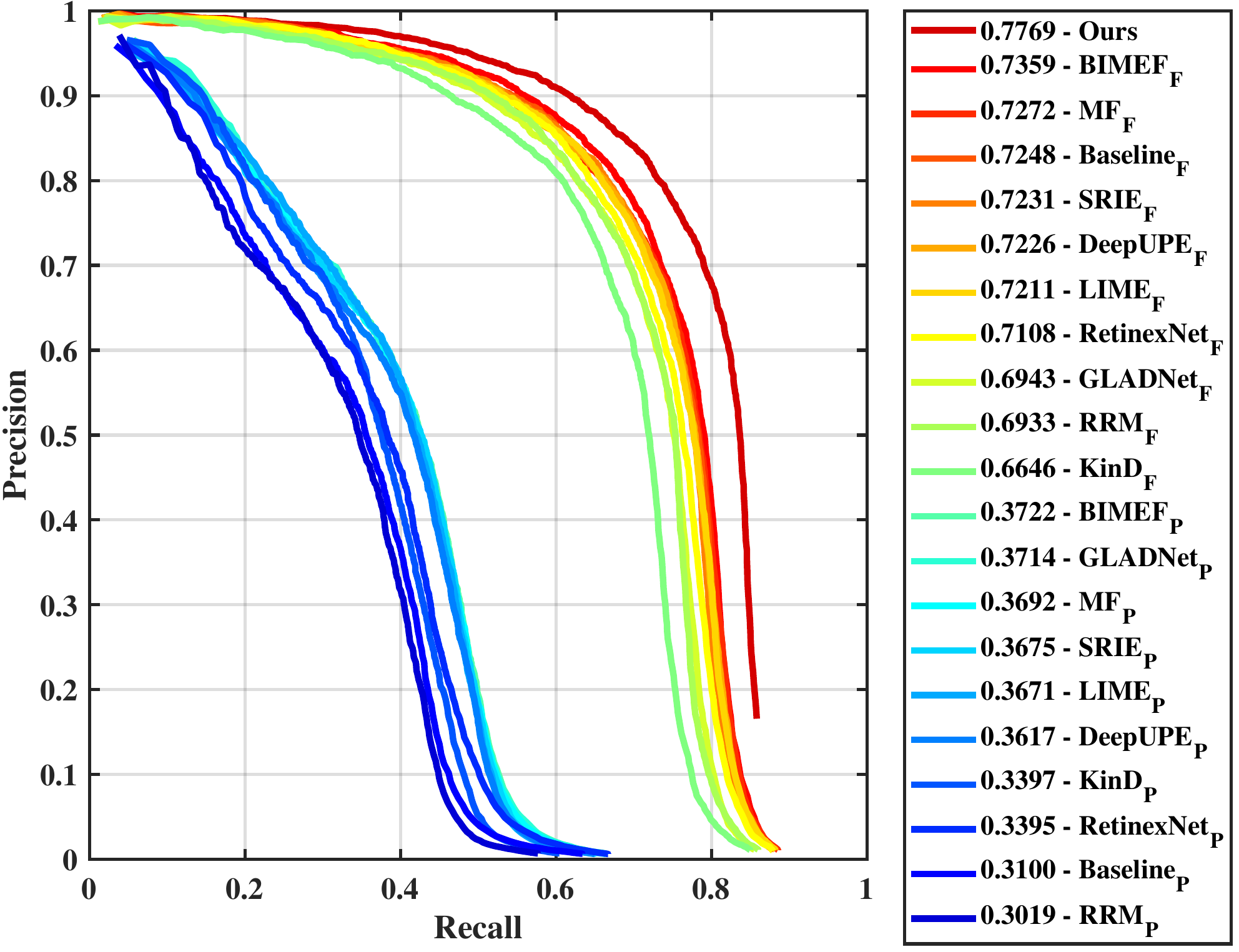}\\
        {\small (b) PyramidBox~\cite{tang2018pyramidbox}} \\
        \includegraphics [width=1.0\linewidth]{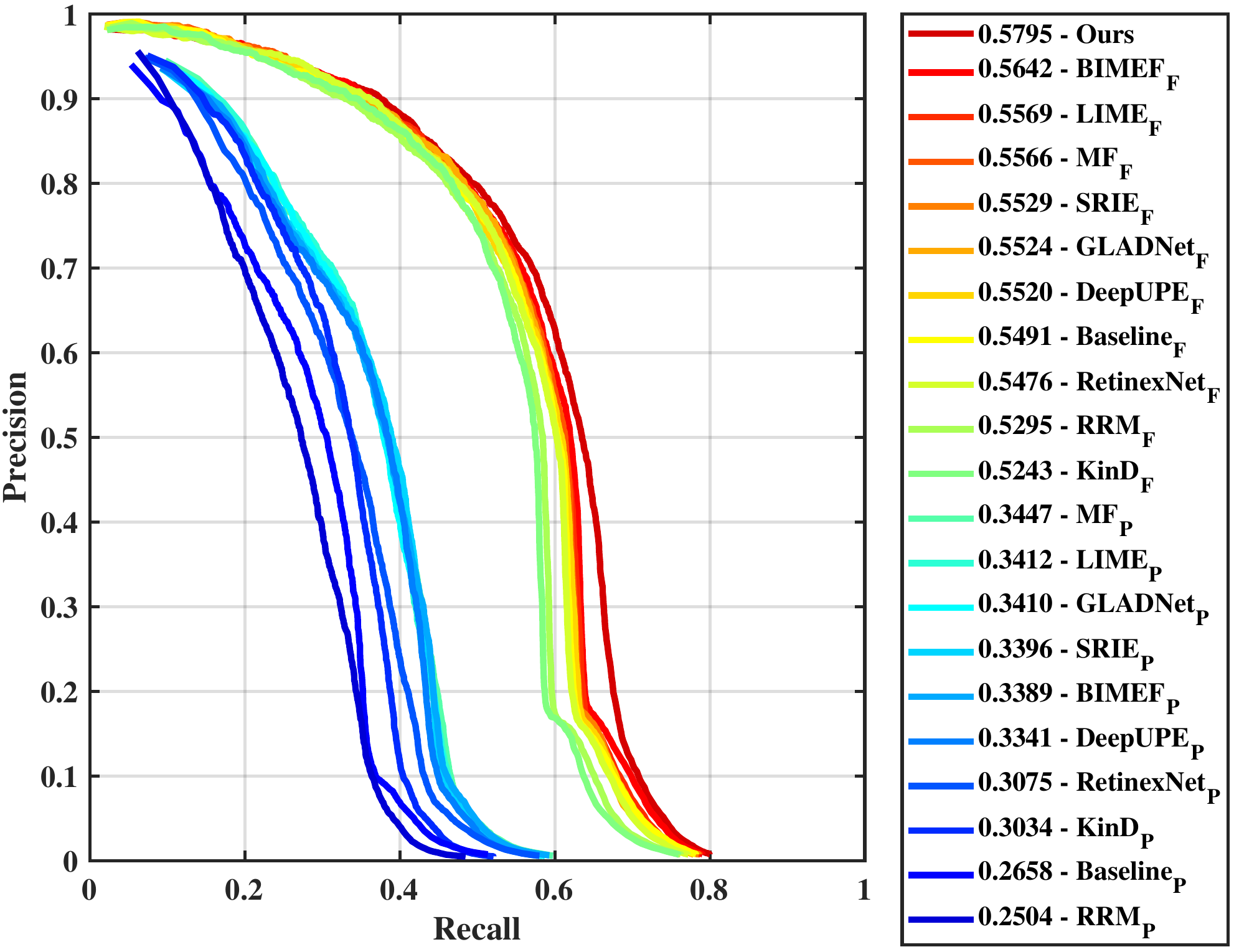}\\
        {\small (c) S3FD~\cite{zhang2017s3fd}}\\
    \end{tabular}
    \caption{
    Quantitative results of different approaches are shown. All the other approaches have both pre-trained version (marked with subscript `P') and finetuned version (marked with subscript `F') excepting for ours.}
    \label{fig:performance}
\end{figure}

\begin{figure*}[htbp!]
\centering
\begin{tabular}{@{}c@{\extracolsep{2pt}}c@{\extracolsep{2pt}}c@{\extracolsep{2pt}}c@{\extracolsep{2pt}}c@{}}
     \rotatebox{270}{\footnotesize Ours} & 
     \includegraphics[align=t,width=0.205\linewidth]{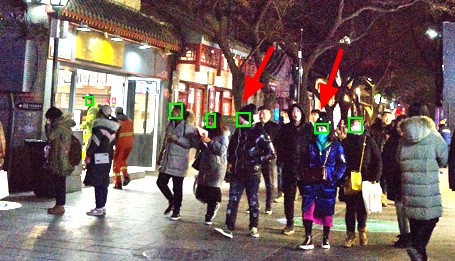} &
     \includegraphics[align=t,width=0.205\linewidth]{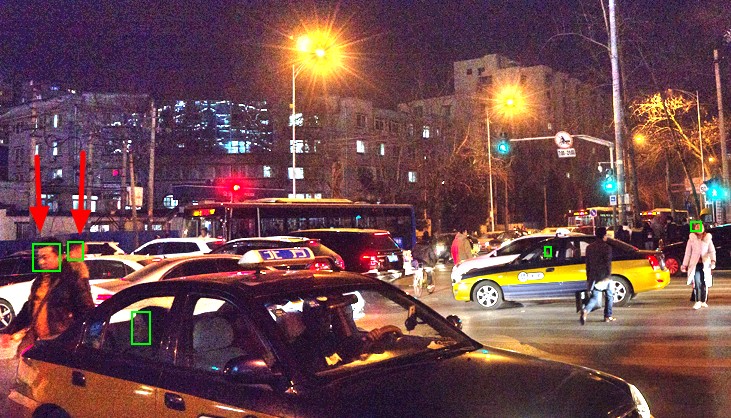} &
     \includegraphics[align=t,width=0.205\linewidth]{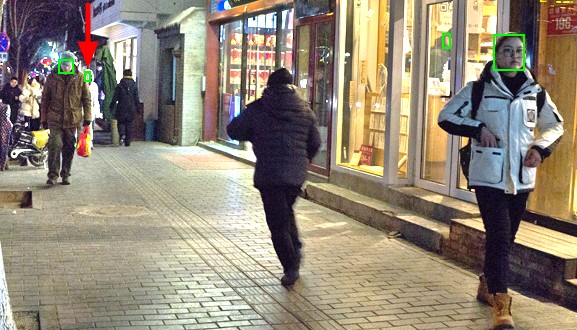} &
     \includegraphics[align=t,width=0.205\linewidth]{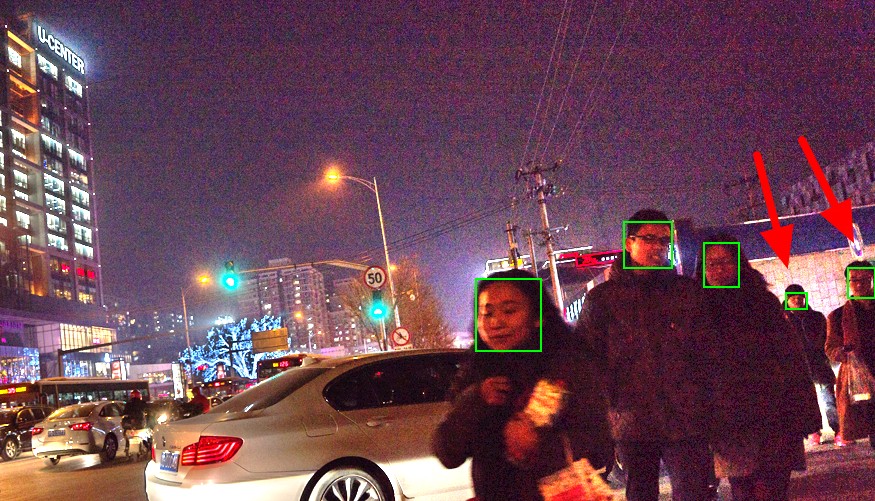} \\
     \rotatebox{270}{\footnotesize KinD~\cite{zhang2019kindling}} & 
     \includegraphics[align=t,width=0.205\linewidth]{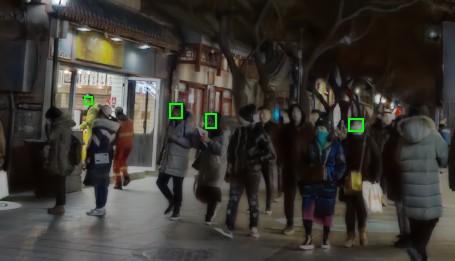} &
     \includegraphics[align=t,width=0.205\linewidth]{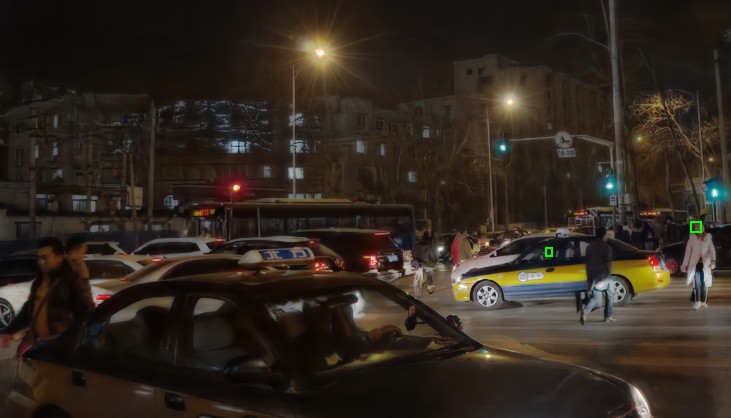} &
     \includegraphics[align=t,width=0.205\linewidth]{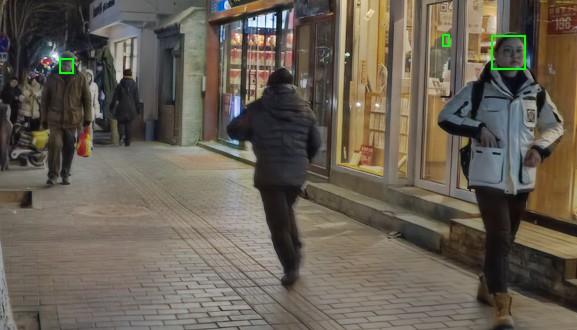} &
     \includegraphics[align=t,width=0.205\linewidth]{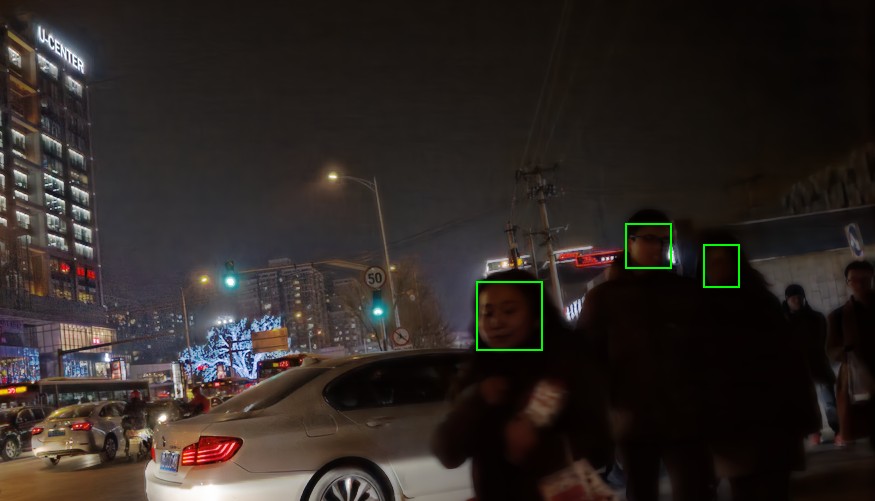} \\
     \rotatebox{270}{\footnotesize DeepUPE~\cite{wang2019underexposed}} & 
     \includegraphics[align=t,width=0.205\linewidth]{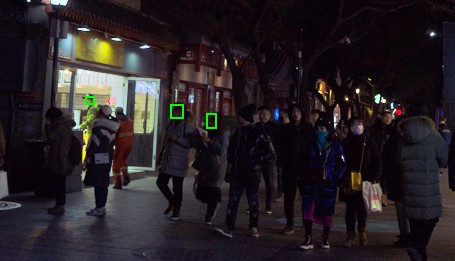} &
     \includegraphics[align=t,width=0.205\linewidth]{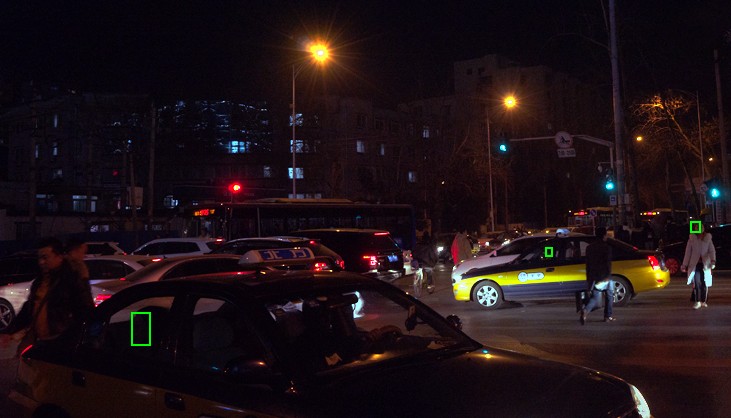} &
     \includegraphics[align=t,width=0.205\linewidth]{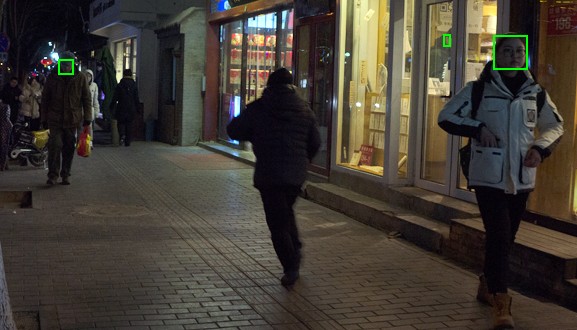} &
     \includegraphics[align=t,width=0.205\linewidth]{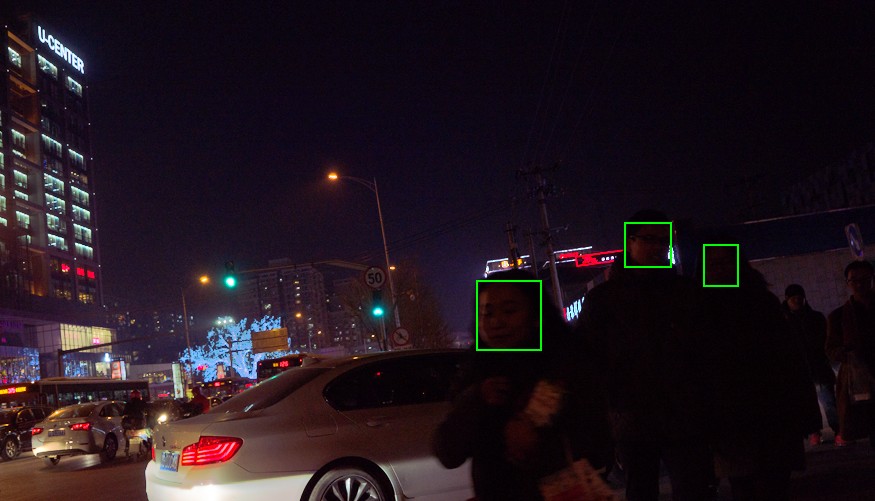} \\
     \rotatebox{270}{\footnotesize RRM~\cite{li2018structurerevealing}} & 
     \includegraphics[align=t,width=0.205\linewidth]{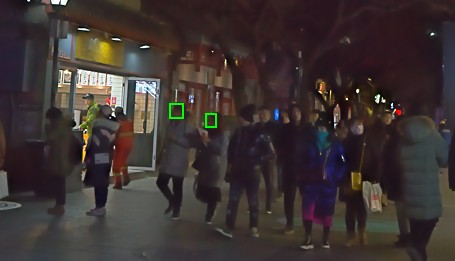} &
     \includegraphics[align=t,width=0.205\linewidth]{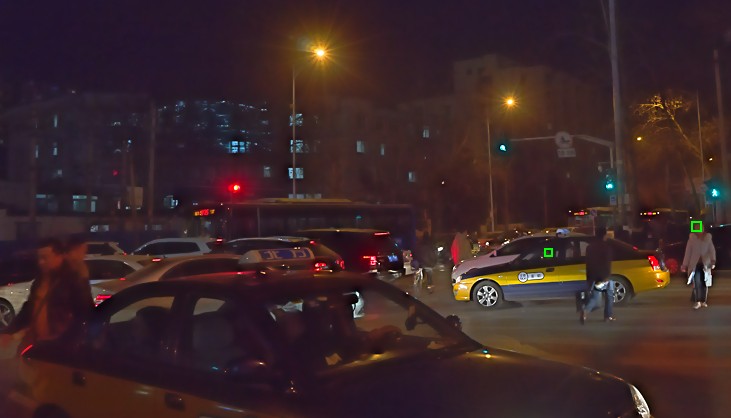} &
     \includegraphics[align=t,width=0.205\linewidth]{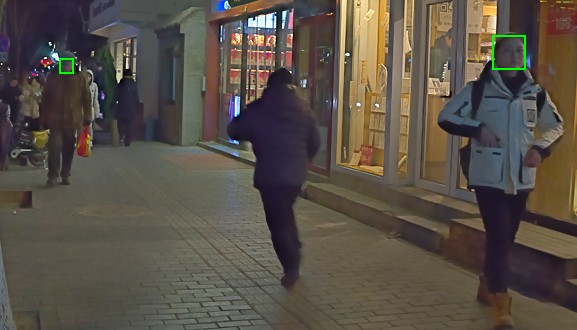} &
     \includegraphics[align=t,width=0.205\linewidth]{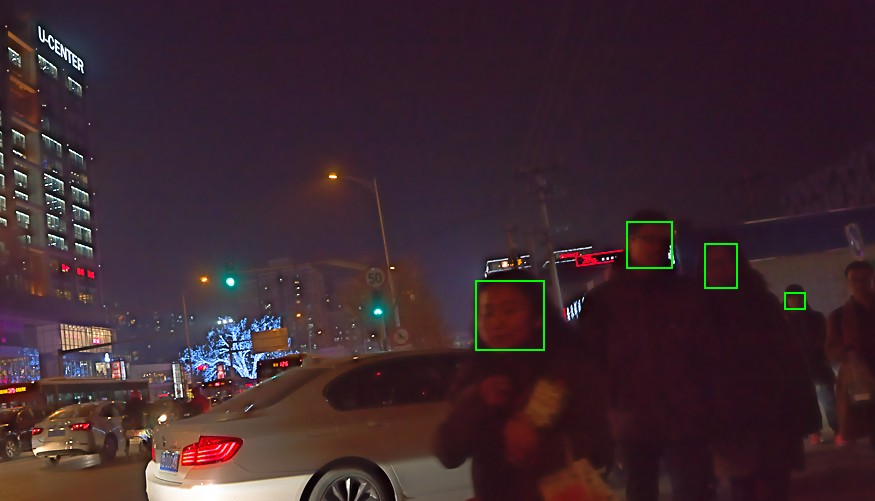} \\
     \rotatebox{270}{\footnotesize RetinexNet~\cite{wei2018deep}} & 
     \includegraphics[align=t,width=0.205\linewidth]{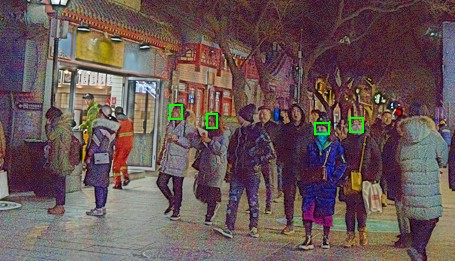} &
     \includegraphics[align=t,width=0.205\linewidth]{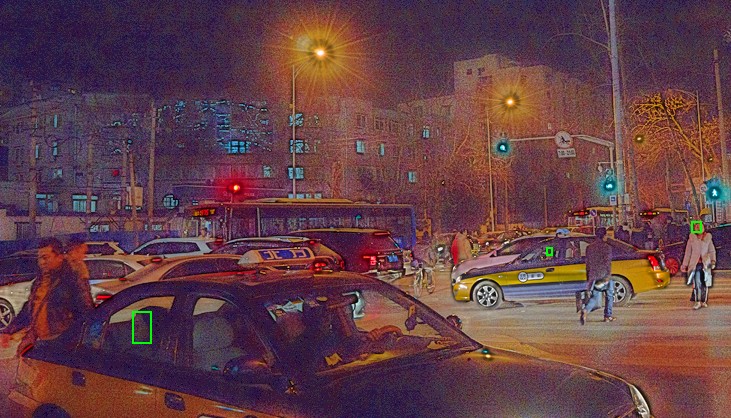} &
     \includegraphics[align=t,width=0.205\linewidth]{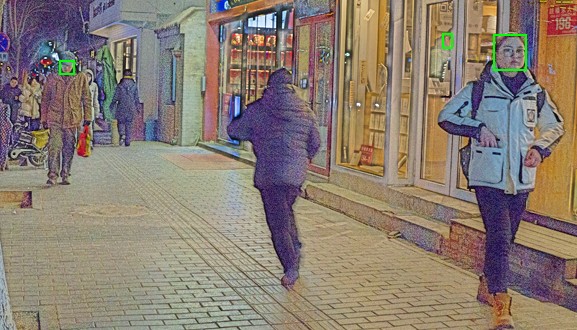} &
     \includegraphics[align=t,width=0.205\linewidth]{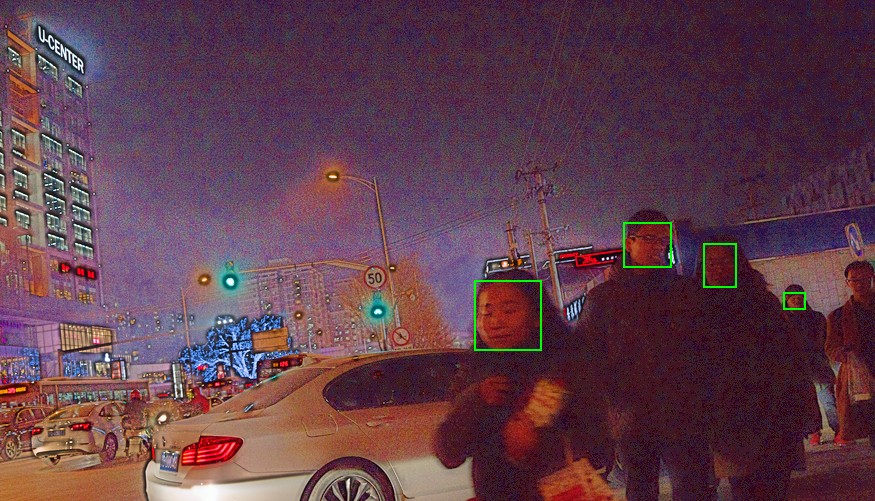} \\
     \rotatebox{270}{\footnotesize GLADNet~\cite{wang2018gladnet}} & 
     \includegraphics[align=t,width=0.205\linewidth]{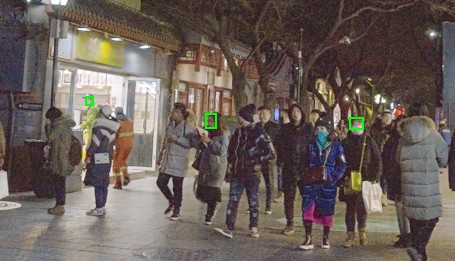} &
     \includegraphics[align=t,width=0.205\linewidth]{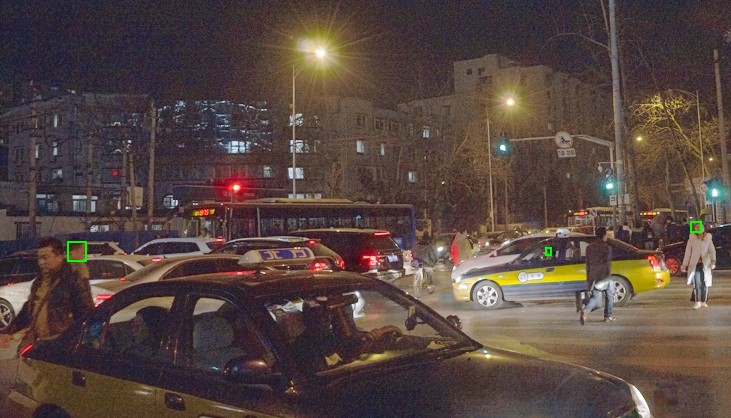} &
     \includegraphics[align=t,width=0.205\linewidth]{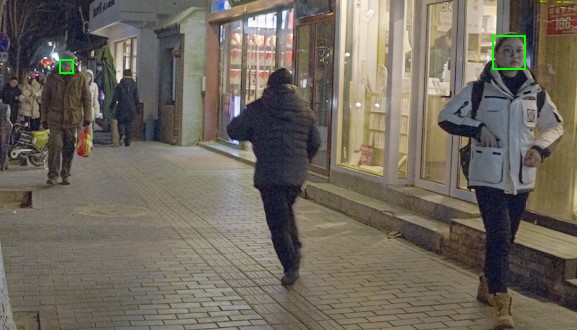} &
     \includegraphics[align=t,width=0.205\linewidth]{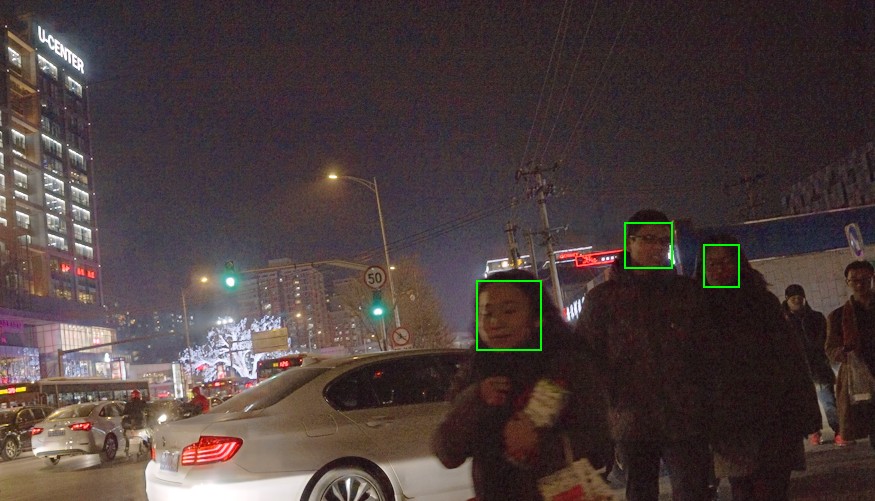} \\
     \rotatebox{270}{\footnotesize LIME~\cite{guo2017lime}} & 
     \includegraphics[align=t,width=0.205\linewidth]{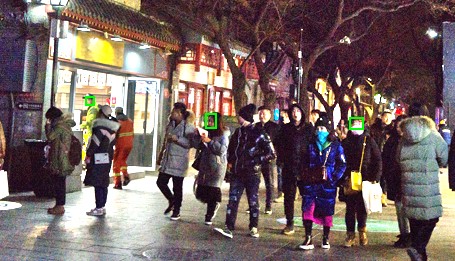} &
     \includegraphics[align=t,width=0.205\linewidth]{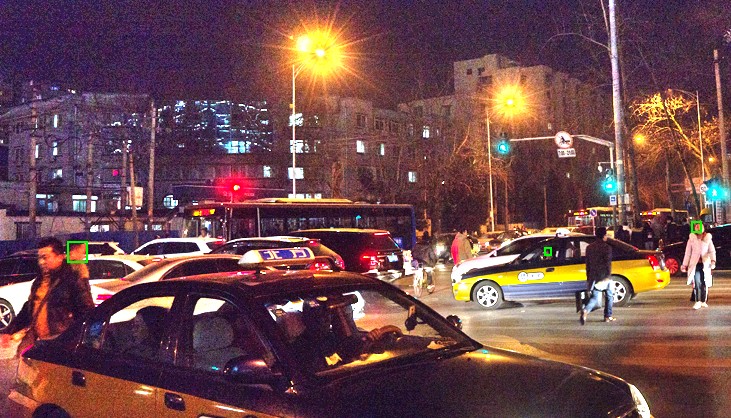} &
     \includegraphics[align=t,width=0.205\linewidth]{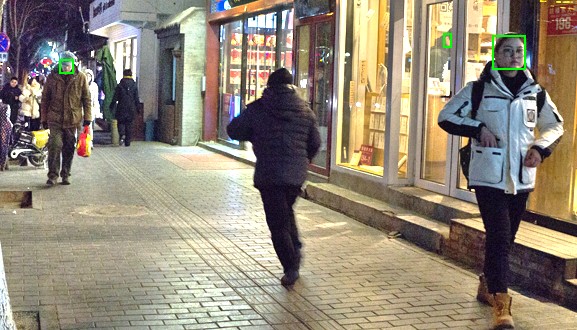} &
     \includegraphics[align=t,width=0.205\linewidth]{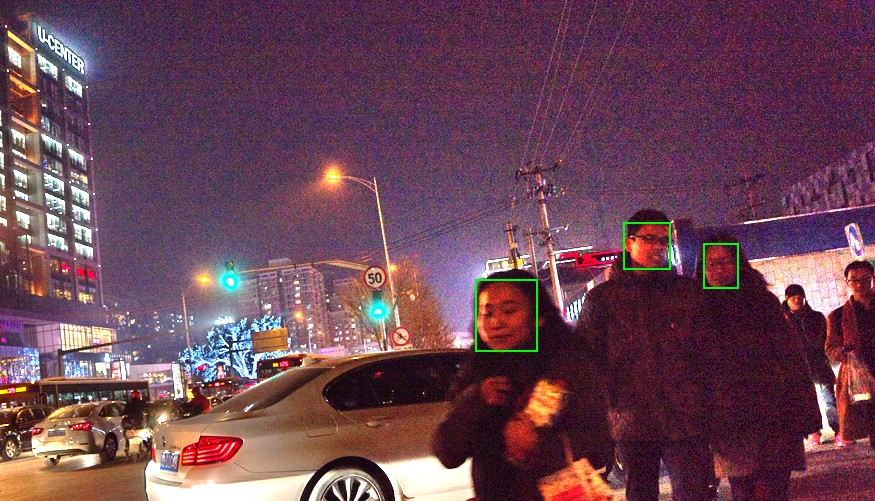} \\
     \rotatebox{270}{\footnotesize BIMEF~\cite{ying2017bioinspired}} & 
     \includegraphics[align=t,width=0.205\linewidth]{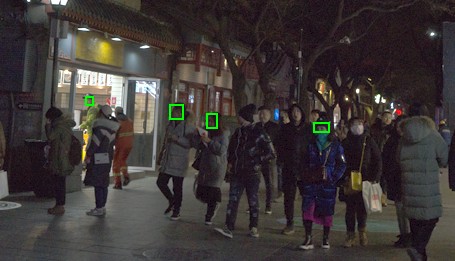} &
     \includegraphics[align=t,width=0.205\linewidth]{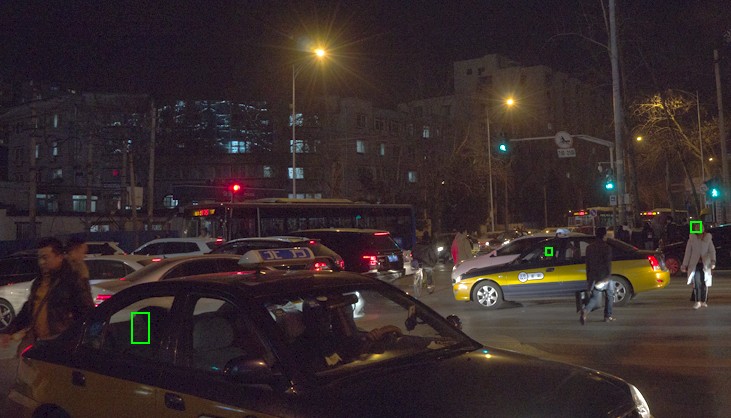} &
     \includegraphics[align=t,width=0.205\linewidth]{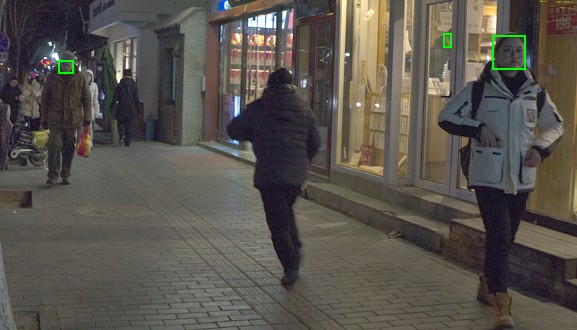} &
     \includegraphics[align=t,width=0.205\linewidth]{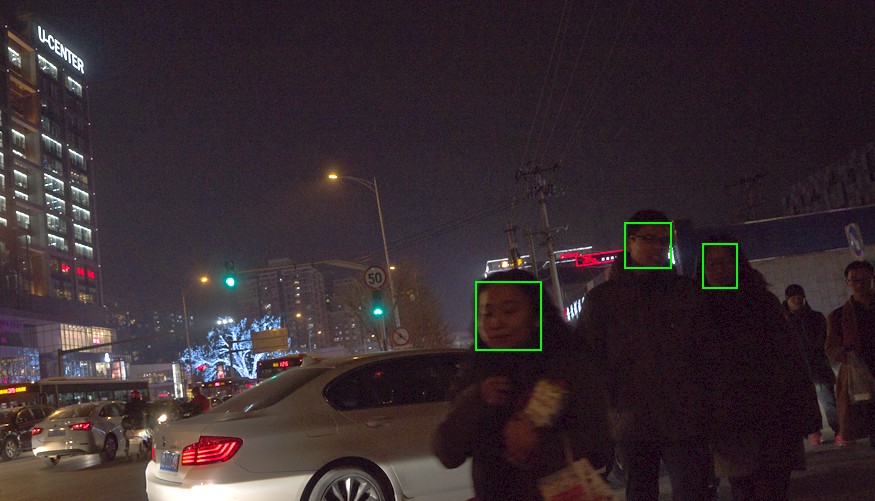} \\
     \rotatebox{270}{\footnotesize SRIE~\cite{fu2016weighted}} & 
     \includegraphics[align=t,width=0.205\linewidth]{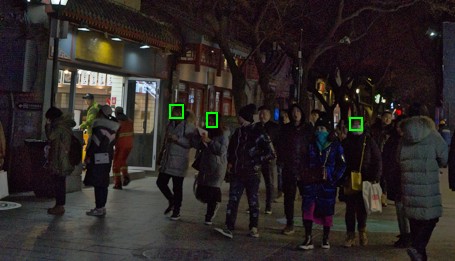} &
     \includegraphics[align=t,width=0.205\linewidth]{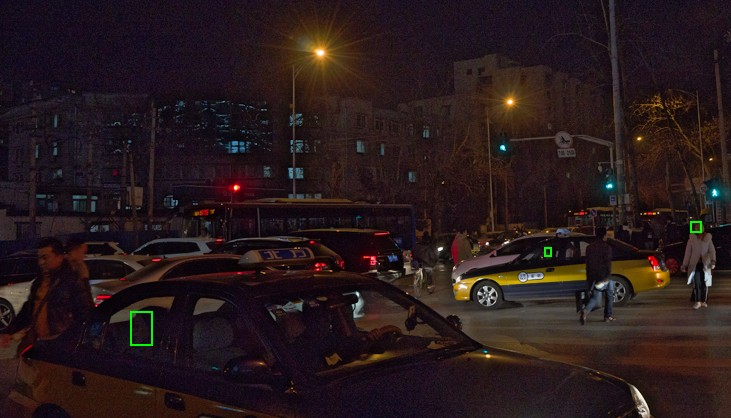} &
     \includegraphics[align=t,width=0.205\linewidth]{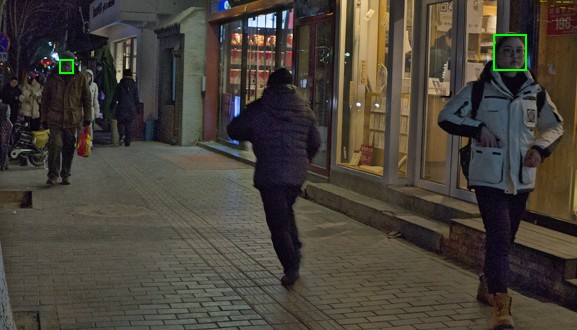} &
     \includegraphics[align=t,width=0.205\linewidth]{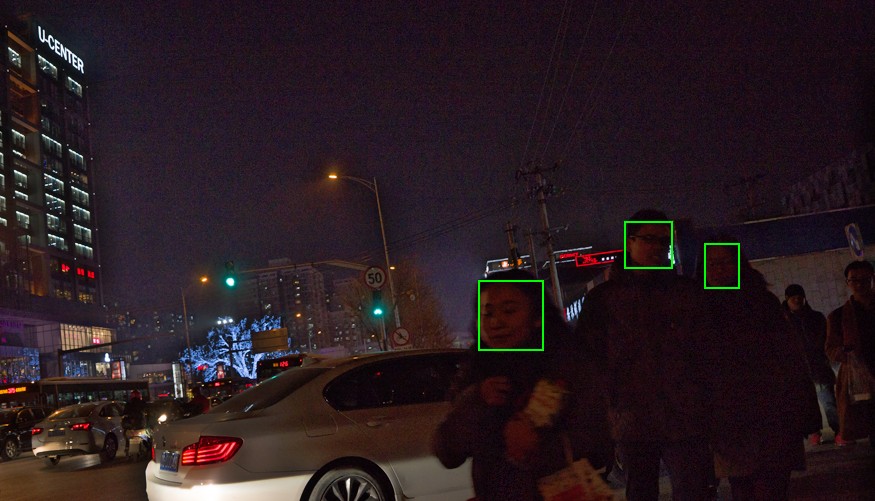} \\
     \rotatebox{270}{\footnotesize MF~\cite{fu2016fusionbased}} & 
     \includegraphics[align=t,width=0.205\linewidth]{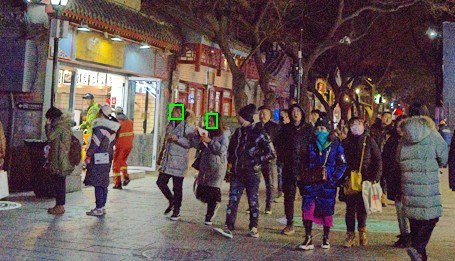} &
     \includegraphics[align=t,width=0.205\linewidth]{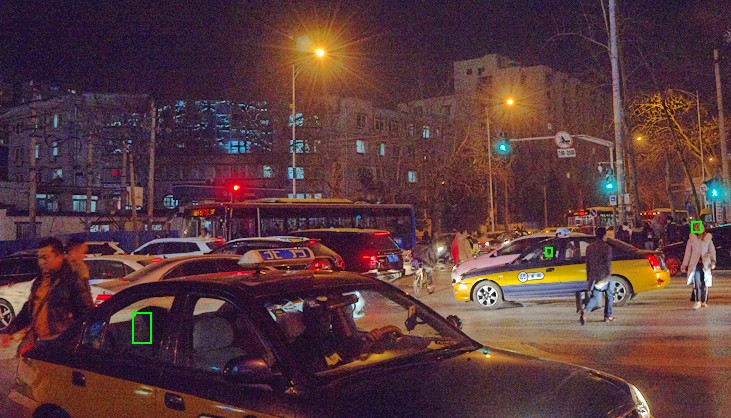} &
     \includegraphics[align=t,width=0.205\linewidth]{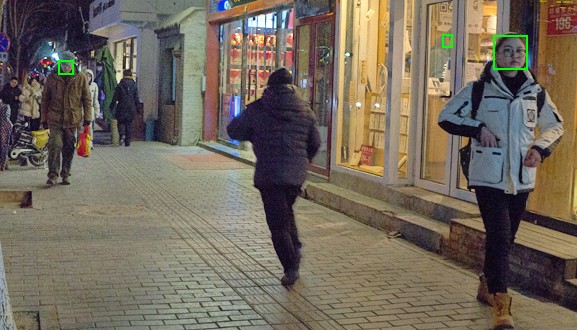} &
     \includegraphics[align=t,width=0.205\linewidth]{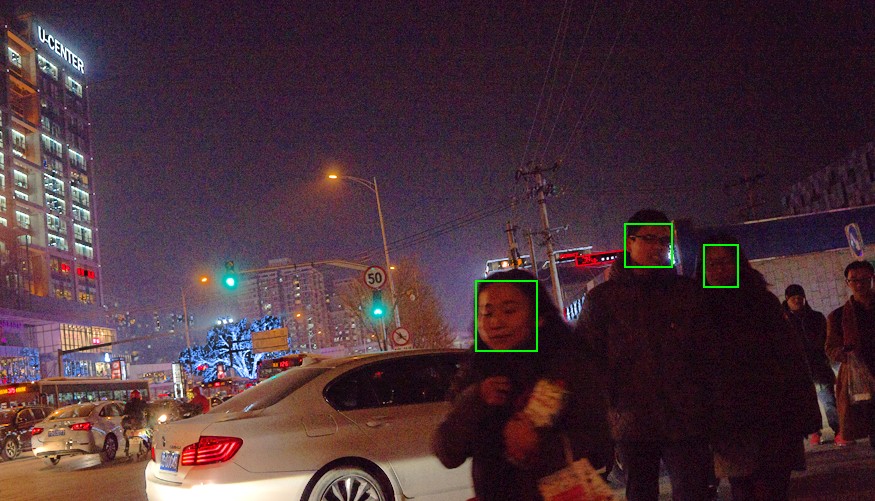} \\
\end{tabular}
    \caption{Qualitative comparison of different methods. For better visualization, we draw the results of REGDet on images enhanced by LIME~\cite{guo2017lime}. \textcolor{red}{Red} arrows indicate those faces that are challenging to be detected by the other methods. \textit{Please Zoom in to see better}.
    }    
    \label{fig:qualitative}
\end{figure*}

\section{Experiments}
\label{sec:experiments}

\subsection{Setup}
\subsubsection{Dataset and metric}
We adopt the recently constructed DARK FACE dataset~\cite{yang2019ug} as our testbed. 6,000 real-world low-light images captured under extreme low-light environment. The resolution of the images is $1080 \times 720$. Totally 43,849 manually annotated faces are released. The annotated faces have large scale variance, ranging from $1 \times 2$ to $335 \times 296$. There are usually $1$ to $20$ annotated faces in an image.
Since the original test split~\cite{yang2019ug} is withheld, we randomly leave 1000 images as our test set. 
Following prior work~\cite{li2019dsfd,tang2018pyramidbox,zhang2017s3fd}, face detection performance is measured by mean Average Precision (mAP), which is calculated as the area under precision-recall curve. 

\subsubsection{Base detectors}
To benefit from the publicly available pre-trained models, we build up REGDet on the \textit{base detectors} pre-trained on the existing largest dataset for face detection in the wild, \ie, WIDER FACE~\cite{yang2016wider} dataset. 
DSFD~\cite{li2019dsfd}, PyramidBox~\cite{tang2018pyramidbox} and S3FD~\cite{zhang2017s3fd}, the state-of-the-art methods that achieve remarkable performance on WIDER FACE, are chosen as the base detectors. 
The weights of REGDet are initialized and bootstrapped as described in Section \ref{sec:reg} and \ref{sec:med}. 
For reproducibility, we adopt public implementation of the base detectors
with VGG-16 backbone network.
As photometric augmentation has been a common practice in modern detectors, data augmentation related to exposure levels is used in the compared baselines.

\subsubsection{Implementation details}
Following~\cite{liu2016ssd,li2019dsfd,tang2018pyramidbox}, the batch size is 16 and multiple GPUs are used for speedup. The initial learning rate is 0.001, which is decreased by 0.1 at the 64-th and 96-th epoch. We adopt Adam~\cite{kingma2015adam} to train the REG module and SGD with momentum of 0.9 to train the MED module. 
Face anchor boxes that have over $0.35$ IoU with the ground-truth annotated faces are labeled as positive anchors. The ratio between sampled negative anchors and positive anchors is fixed to $3$ at each training iteration. For our proposed REGDet, we remove random photometric distortion in data augmentation as it has already involved an enhancement module. Note that we keep the photometric augmentation for the baselines following~\cite{li2019dsfd,tang2018pyramidbox, zhang2017s3fd} for fair comparison. 
During inference, the image is first rescaled to make $\sqrt{H \times W}=2000$. 
Non-maximum suppression is applied with Jaccard overlap of 0.3 and the top 750 bounding boxes are kept.

\subsubsection{Compared methods}
We compare REGDet against various face detectors with illumination pre-processing using the state-of-the-art low-light image enhancement approaches
including \textbf{MF}~\cite{fu2016fusionbased}, \textbf{SRIE}~\cite{fu2016weighted}, \textbf{LIME}~\cite{guo2017lime},  \textbf{BIMEF}~\cite{ying2017bioinspired},  \textbf{GLADNet}~\cite{wang2018gladnet}, \textbf{RetinexNet}~\cite{wei2018deep}, \textbf{RRM}~\cite{li2018structurerevealing},  \textbf{DeepUPE}~\cite{wang2019underexposed}, and \textbf{KinD}~\cite{zhang2019kindling}
to preprocess the images. 
\textbf{Baseline} denotes the plain detector fed by the original low-light images as input.
We evaluate all the aforementioned approaches with both pre-trained and finetuned version. The pre-trained version directly uses the pre-trained weights on WIDER FACE and performs inference on pre-processed DARK FACE images using the aforementioned methods. The finetuned version further finetunes the model using pre-processed DARK FACE images as input. As the performances reported in \cite{yang2019ug} are for the withheld test data split with only pre-trained version, \textit{we re-train the aforementioned methods on our train split and fairly compare them on our 1000-image test split}.

\begin{figure*}[htbp!]
    \centering
    \begin{tabular}{@{}c@{\extracolsep{2pt}}c@{\extracolsep{2pt}}c@{\extracolsep{2pt}}c@{}c@{}}
    \includegraphics[width=0.23\linewidth]{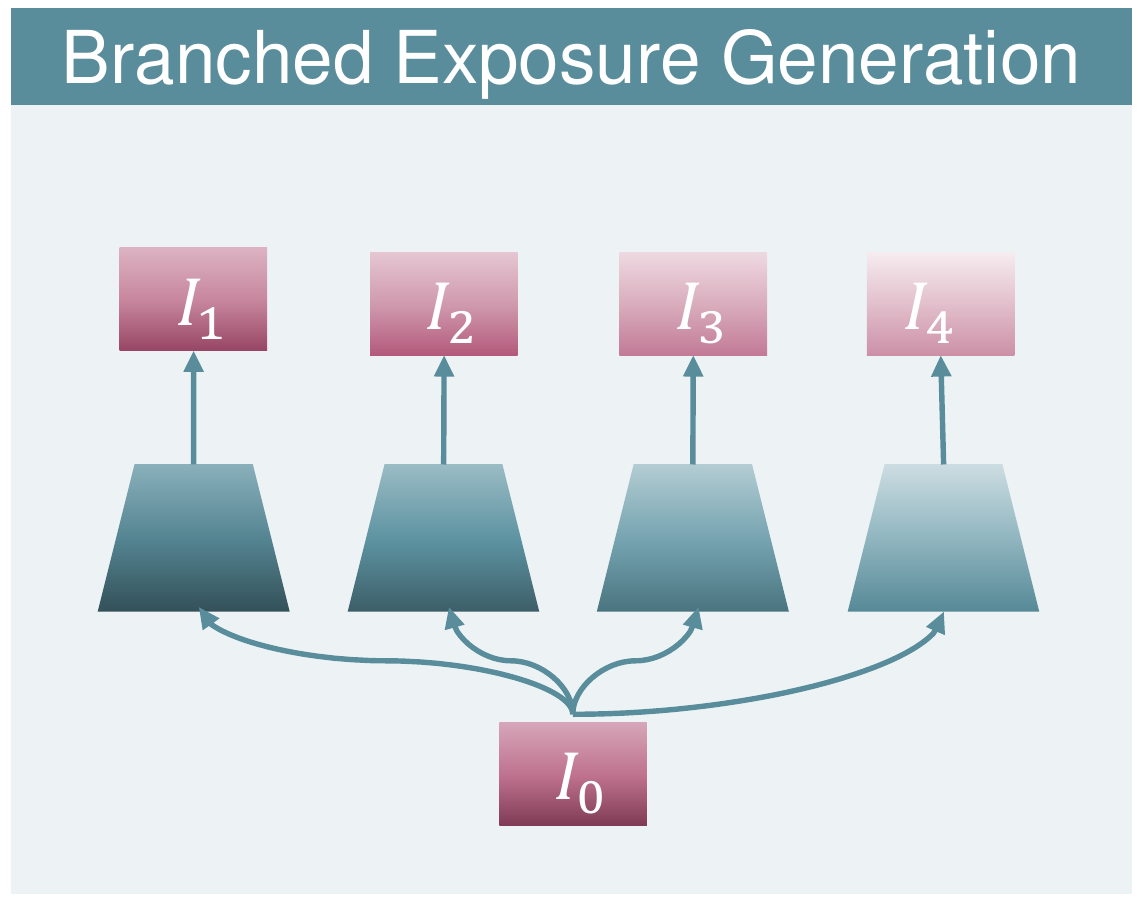} & 
    \includegraphics[width=0.23\linewidth]{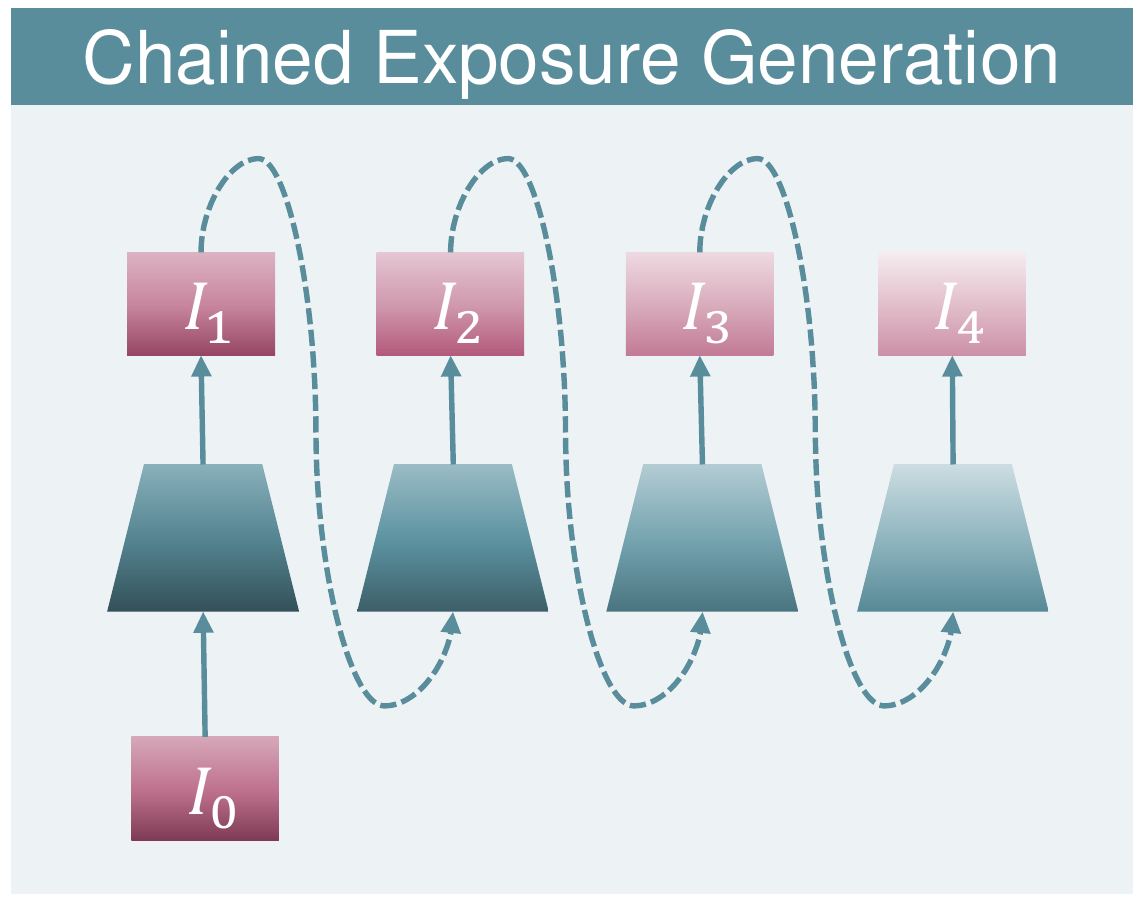} & 
    \includegraphics[width=0.23\linewidth]{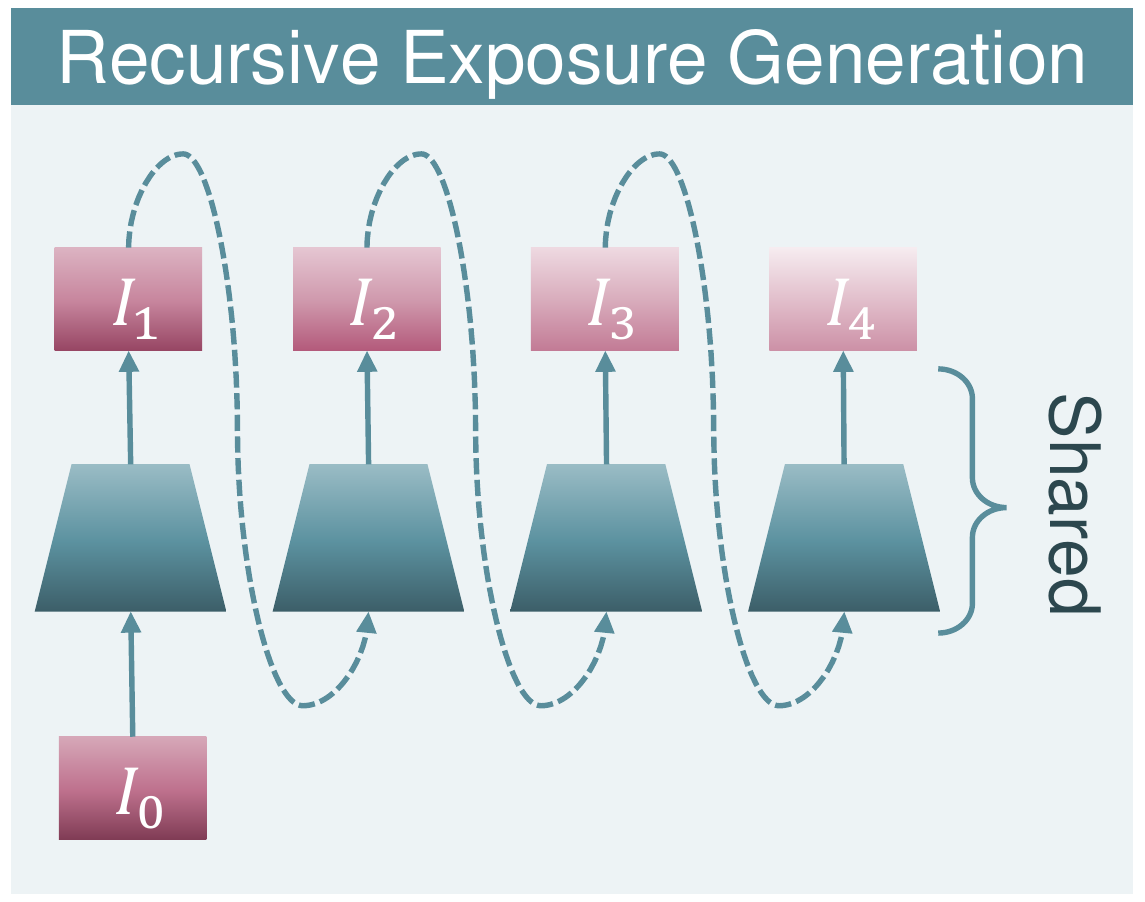} & 
    \includegraphics[width=0.23\linewidth]{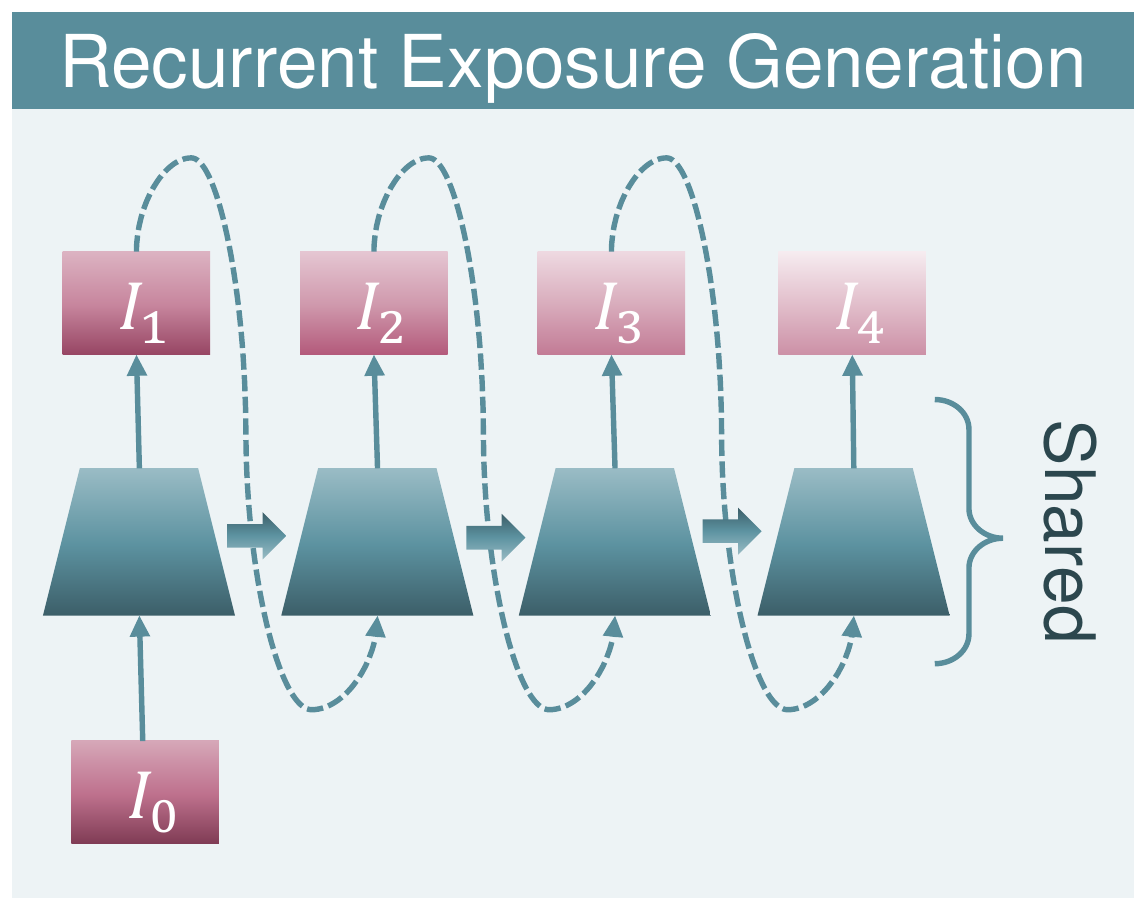} & 
    \includegraphics[width=0.07\linewidth]{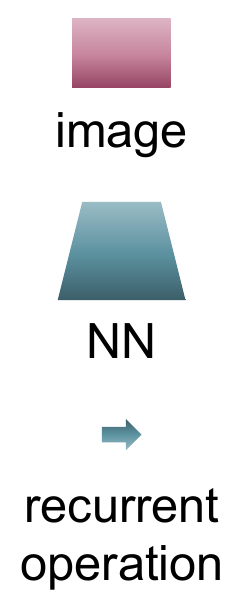} \\
    {\small (a) BEG} & {\small (b) CEG} & {\small (c) SEG} & {\small (d) REG} & 
    \end{tabular}
    \caption{Alternative pseudo-exposure generation modules.}
    \label{fig:enhance}
\end{figure*}

\begin{table*}[htbp!]
 \centering
  \caption{Results of ablation study on the proposed REG module.} %
  \label{tab:ablation}
    \begin{tabular}{crrrrrr}
    \toprule
    \multirow{2}{*}{Method}&
    \multicolumn{2}{c}{DSFD~\cite{li2019dsfd}}&\multicolumn{2}{c}{PyramidBox~\cite{tang2018pyramidbox}}&\multicolumn{2}{c}{S3FD~\cite{zhang2017s3fd}}\cr
    \cmidrule(lr){2-3} \cmidrule(lr){4-5} \cmidrule(lr){6-7}
    & \#Params & mAP (\%) & \#Params & mAP (\%) & \#Params & mAP (\%)\cr
    \midrule
    Finetuned Baseline
    & 47.49M & 71.42 
    & 54.53M & 72.48  
    & 21.42M & 54.99 \cr
    Ours-BEG 
    & + 0.09M & 75.60
    & + 0.09M & 76.11 
    & + 0.09M & 56.78\cr
    Ours-CEG 
    & + 0.09M & 74.07 
    & + 0.09M & 73.16
    & + 0.09M & 54.30\cr
    Ours-SEG 
    & + 0.03M & 73.52
    & + 0.03M & 74.19 
    & + 0.03M & 52.82\cr
    Ours-REG 
    & + 0.12M & \textbf{76.94} 
    & + 0.12M & \textbf{77.69}
    & + 0.12M & \textbf{57.95}\cr
    \bottomrule
    \end{tabular}
\end{table*}

\subsection{Result Analysis}
The quantitative comparison of different approaches is shown in Fig.~\ref{fig:performance}. 
The three pre-trained baseline detectors achieve results of 32.69\%, 31.00\%, and 26.58\% mAP respectively. The relative performance disparity among the three detectors are consistent with their performance on  WIDER FACE. 
The former two detectors perform much better as they apply modern context aggregation techniques such as feature enhancement using two shots~\cite{li2019dsfd} or context assisted pyramid anchors~\cite{tang2018pyramidbox}. Compared with the pre-trained detectors, all finetuned ones achieve much higher performance, indicating that the existing large-scale dataset WIDER FACE dominated by normal-light images carries very different lighting distribution compared to DARK FACE dataset. 
Compared with original image input, many of the image enhancement approaches improve the face detection performance. 
Specifically, the pre-trained detectors equipped with pre-processing using MF, LIME, BIMEF, DeepUPE, GLADNet, and SRIE outperform the baseline with respectively 4.87\%, 5.08\%, 5.33\%, 4.60\%, and 0.45\% performance gain when using DSFD as the base detector. 
In the finetuned setting, MF, LIME, BIMEF, and DeepUPE improve the baseline with respectively 1.12\%, 0.94\%, 1.75\%, and 1.05\% performance gain when using DSFD as the base detector. 
While these image enhancement methods show clear advantages over the baseline with the pre-trained setting, they achieve less performance gain in the finetuned setting, as finetuning already greatly reduces the data distribution discrepancy between normal-light and low-light images.
However, it is noticeable that KinD, RetinexNet, and RRM cause performance degeneration to different extents due probably to the severe over-smoothness (KinD, RRM) or artifacts (RetinexNet) on regions containing faces (also evidenced by Fig.~\ref{fig:qualitative}. 
Among them, the multi-exposure fusion method BIMEF performs best. 
The relatively good performance of BIMEF may also imply that it is promising to adaptively generate pseudo exposures with different light conditions, which is consistent with what we explored in this paper.
In particular, compared with the finetuned baseline on original images equipped with photometric data augmentation~\cite{howard2013improvements}, the proposed REGDet shows much higher detection mAP with respectively about \textbf{5.5\%}, \textbf{5.2\%}, and \textbf{3.0\%} performance gain using the three base detectors, with negligible extra parameters (as shown in Table~\ref{tab:ablation}). 
The overwhelmingly high detection rates of REGDet demonstrates its superiority over existing state-of-the-arts.

The qualitative results of different approaches on sampled images from DARK FACE are shown in Fig.~\ref{fig:qualitative}. 
While those large and clear faces can also be detected by other methods, our method has successfully found much more dark and tiny faces, as pointed out by the red arrows in the presented images. Although it is hard to detect those faces even by human eyes, the proposed method is able to localize most of them and clearly outperforms other approaches.

\subsection{Ablation Studies}


\subsubsection{Effectiveness of the recurrent architecture}
To examine the effectiveness of the proposed recurrent component, variant generation modules are designed as illustrated in Fig.~\ref{fig:enhance}, which includes 
\begin{itemize}
\item \textbf{Branched Exposure Generation (BEG)} \quad This module generates different exposures $I_t$ parallelly from the original image $I_0$ by a module with $T$ branches, 
\item \textbf{Chained Exposure Generation (CEG)} \quad The $t$-th image is generated at the $t$-th stage of the module with non-shared weights conditioned on the image $I_{t-1}$ generated at the $(t-1)$-th stage, 
\item \textbf{RecurSive Exposure Generation (SEG)} \quad Similar with CEG, except that the module shares parameters at different stages, 
\item \textbf{Recurrent Exposure Generation (REG)} \quad The module used in our proposed method. Different from the aforementioned modules, REG encodes historical feature maps in order to alleviate the probable unrecoverable information loss caused by the over-exposure and over-smoothness at the middle stages. 
The detailed description of the REG module is provided in Sec.~\ref{sec:method}.
\end{itemize}
 
We replace REG with BEG, CEG, SEG respectively and conduct experiments on DARK FACE. 
As shown in Table~\ref{tab:ablation}, 
all the designed lightweight modules introduce merely a few extra parameters while they almost all achieve improved detection results. 
BEG constructs multiple branches from the original image $I_0$ to generate different pseudo-exposures in parallel, and clearly boosts performance, indicating that the MED module does provide important guidance to the enhancement module for generating complementary information in different pseudo-exposures, as illustrated in Sec.~\ref{sec:med}. 
In contrast, CEG and SEG that generate $I_t$ conditioned on $I_{t-1}$ with  non-shared and shared weight, respectively, produce not so stable performance gain, due probably to unrecoverable information loss caused by the over-exposure and over-smoothness at the middle stages. This suggests that a proper modeling of the multi-exposure generation is the key to achieve good face detection performance. For the performance of using S3FD as base detector, Ours-CEG and Ours-SEG only achieve comparable or even decreased detection rates. We conjecture that the reason of the inferior performance is that S3FD has much less parameters and consequently much smaller model capacity compared with DSFD and PyramidBox, resulting in insufficient guidance effects for the generation modules.
By encoding historical feature maps, the proposed REG alleviates the issue and performs the best. It indicates 
that the relationship between adjacent pseudo-exposures could be well modeled by maintained memory in the recurrent structure of REG.
The consistent performance boost also demonstrates the scalability of REG across different base face detectors. 

\subsubsection{Comparison of different numbers of stages}

We provide experimental comparison of different numbers of stages (denoted as $T$, Section \ref{sec:method}) using PyramidBox as base detector to support our choice of $T = 4$. The results corresponding to $T = 1,2,4,6$ are shown in Table~\ref{tab:ablation_stage}. Setting $T = 1$ is equivalent to a special case of REGDet, namely, a single-exposure `detection-with-enhancement' model. It achieves much higher detection performance (mAP) than the finetuned baseline (72.48\%), but achieves inferior result than the multi-exposure frameworks ($T = 2, 4, 6$).  On one hand, it supports the claim that jointly performing enhancement and detection is superior compared to plain detection for low-light face detection.  On the other hand, it verifies the superiority of the proposed multi-exposure framework over single-exposure framework. Setting $T = 4$ achieves the best performance, indicating that it is a good practice. Setting $T$ to a higher number (e.g., 6) does not bring more performance gain, meaning that generating too many pseudo exposures is unnecessary. Also, setting large $T$ brings heavier computational cost. 

\begin{table}[htbp!]
\centering
    \vspace{10pt}
    \caption{Results of ablation study on the number of stages.}
    \begin{center}
    \begin{tabular}{cc}
    \toprule
       Numbers of stages    &   mAP (\%)   \\
    \midrule
       $T = 1$   &   76.73   \\
       $T = 2$   &   77.16   \\
       $T = 4$   &   \textbf{77.69}   \\
       $T = 6$   &   77.63   \\
    \bottomrule
    \end{tabular}
    \end{center}
    \label{tab:ablation_stage}
\end{table}

\subsubsection{Effectiveness of the pseudo-supervised pre-training}

We provide experimental comparison on whether applying the proposed pseudo-supervised pre-training on the REG module or not. The performance of the resulted REGDet using PyramidBox as base detector are compared in Table~\ref{tab:ablation_pretraining}. When randomly initializing the REG module (w/o pre-training), the proposed REGDet remains good performance with an mAP of 76.36\%. 
Equipped with the proposed pseudo-supervised pre-training technique, our method achieves the best performance with 1.33\% absolute performance gain. 
As illustrated in Sec.~\ref{sec:reg}, the REG module is supervised and guided to generate images corresponding to diversified exposures with the designed pseudo-supervised pre-training. The collaborative and complementary information from different pseudo exposures can potentially be learnt from such a pre-training technique, which we believe is the key for better performance.

\begin{table}[htbp!]
\centering
    \caption{Ablation of the pseudo-supervised pre-training process for the REG module.}
    \begin{center}
    \begin{tabular}{cc}
    \toprule
       Setting    &   mAP (\%)   \\
    \midrule
       Ours w/o pre-training   &   76.36   \\
       Ours w/ pre-training   &   \textbf{77.69}   \\
    \bottomrule
    \end{tabular}
    \end{center}
    \label{tab:ablation_pretraining}
\end{table}

\section{Conclusion}

In this work we proposed an end-to-end face detection framework, named REGDet, for dealing with low-light input images. The key component in REGDet is a novel recurrent exposure generation (REG) module that extends ConvGRU to mimic the multi-exposure technique used in photography. The REG module is then sequentially connected with a multi-exposure detection (MED) module for detecting faces from images under poor lighting conditions. The proposed method significantly outperforms previous algorithms on a public low-light face dataset, with detailed ablation study further validating the  effectiveness of the proposed learning component. 



\ifCLASSOPTIONcaptionsoff
\newpage
\fi


\bibliographystyle{ieee}


\end{document}